\documentclass[12pt]{article}
\usepackage[utf8]{inputenc}
\usepackage{authblk}
\usepackage{amsthm}
\usepackage[labelfont=bf]{caption}

\usepackage{geometry}
\usepackage{amsmath}
\usepackage{amsfonts}
\usepackage{amssymb}
\usepackage{graphicx}
\usepackage{booktabs} 
\usepackage{algorithm}
\usepackage{algorithmic}
\usepackage{hyperref}
\usepackage{multirow}
\usepackage{subscript}
\usepackage{enumitem}
\usepackage[authoryear]{natbib}
\allowdisplaybreaks
\providecommand{\keywords}[1]{\textbf{Keywords} #1}

\newcommand{\E}{\mathbb{E}}
\newcommand{\V}{\mathbb{V}}
\newcommand{\cov}[2]{\mathrm{cov}[#1,#2]}
\newcommand{\T}{\top}
\newcommand{\D}{\mathcal{D}}
\newcommand{\N}{\mathsf{N}}
\newcommand{\GP}[2]{\mathsf{GP}(#1,#2)}
\newcommand{\bs}[1]{\boldsymbol{#1}}
\newcommand{\bb}[1]{\mathbb{#1}}
\newcommand{\xb}{\bs x}

\newcommand{\init}{\mathrm{init}}
\newcommand{\next}{\mathrm{next}}
\newcommand{\glob}{\mathrm{global}}
\newcommand{\ninit}{n_{\init}}
\newcommand{\nmax}{n_{\mathrm{max}}}
\newcommand{\fmin}{f_{\mathrm{min}}}
\newcommand{\fnext}{y_{\next}}
\newcommand{\fglob}{f_{\glob}}
\newcommand{\xglob}{\xb_{\glob}}
\newcommand{\xnext}{\xb_{\next}}

\newcommand{\e}{\varepsilon}
\newcommand{\argmin}{\operatornamewithlimits{argmin}}
\newcommand{\argmax}{\operatornamewithlimits{argmax}}
\newcommand{\fhat}{\hat{f}}
\renewcommand{\d}{\operatorname{d}}
\newcommand{\tabularnewline}{\\}

\begin{document}

\title{On a New Improvement-Based Acquisition Function for Bayesian Optimization}

\author[1]{Umberto No\`{e}\thanks{\texttt{umberto.noe@dzne.de}}}
\author[2]{Dirk Husmeier}

\affil[1]{German Center for Neurodegenerative Diseases (DZNE),
Bonn, Germany}
\affil[2]{School of Mathematics
and Statistics, University of Glasgow, Glasgow, United Kingdom}

\date{}

\maketitle

\begin{abstract}
Bayesian optimization (BO) is a popular algorithm for solving challenging
optimization tasks. It is designed for problems where the objective
function is expensive to evaluate, perhaps not available in exact
form, without gradient information and possibly returning noisy values.
Different versions of the algorithm vary in the choice of the acquisition
function, which recommends the point to query the objective at next.
Initially, researchers focused on improvement-based acquisitions,
while recently the attention has shifted to more computationally expensive
information-theoretical measures. In this paper we present two major
contributions to the literature. First, we propose a new improvement-based
acquisition function that recommends query points where the improvement
is expected to be high with high confidence. The proposed algorithm
is evaluated on a large set of benchmark functions from the global
optimization literature, where it turns out to perform at least as
well as current state-of-the-art acquisition functions, and often
better. This suggests that it is a powerful default choice for BO.
The novel policy is then compared to widely used global optimization
solvers in order to confirm that BO methods reduce the computational
costs of the optimization by keeping the number of function evaluations
small. The second main contribution represents an application 
to precision medicine, where the interest lies in the estimation of
parameters of a partial differential equations model of the human
pulmonary blood circulation system. Once inferred, these parameters can help clinicians
in diagnosing a patient with pulmonary hypertension without going
through the standard invasive procedure of right heart catheterization,
which can lead to side effects and complications (e.g. severe pain, internal bleeding, thrombosis). 
\end{abstract}

\newpage
\section{Introduction\label{sec:introduction}}

Many real-life applications such as parameter estimation and decision
making in science, engineering and economics require solving an optimization
problem. Traditionally the aim has been to minimize a function which
is fast to query at any given point and where the gradient information
is readily available or easy to estimate. More recently, new research
directions involve complex and multiscale computational models that
do not meet these requirements. They typically are computationally
expensive, perhaps without an exact functional form, the gradient
information might not be available and outputs could be corrupted
by noise. Examples of these applications include parameter estimation
in robotics \citep{Calandra2016,Lizotte2007}, automatic tuning of
machine learning algorithms \citep{Hutter2011,Snoek2012,Wang2013,Kotthoff2017},
environmental monitoring and sensor placement \citep{Garnett2010},
soft-tissue mechanical models of the pulmonary circulation \citep{Noe2017}
and more~\citep{Shahriari2016}.

Bayesian optimization (BO) is a class of algorithms designed to solve
these complex optimization tasks. It is not a recent field as it dates
back to the 1970\textendash 1980s, when the Lithuanian mathematician
Jonas Mockus published a series of papers and a book on the topic~\citep{Mockus1975,Mockus1977,Mockus1989},
but it increased in popularity only recently due to the advances in
computational resources.

With the increase in computational power, modellers started developing
more complex \emph{simulators} of real life phenomena. For example,
by switching from linear to nonlinear differential equations, adding
more layers of them, and interfacing many micro-level models in order
to recreate in silico macro-level phenomena. Simulators typically
involve many tunable parameters. Setting these parameters by hand
would be cumbersome, hence the need for an automated and principled
framework to deal with them. In these models standard likelihood based
inference is not straightforward due to the time required for a single
forward simulation. This could involve, for example, the numerical
solution of a system of nonlinear partial differential equations,
hence calling for an iterative procedure. Furthermore, the maximum
likelihood equations may not have an analytical solution and need
to be solved iteratively, adding another level of computational complexity
to the problem. Usually the likelihood landscape is highly multimodal,
calling for multiple restarts, and effectively making the problem
NP-hard.

Section~\ref{sec:background} states the generic problem and reviews
the main aspects of Gaussian processes and Bayesian optimization.
It then summarizes the popular classes of acquisition functions found
in the literature, emphasizing the line of thought that led to their
development. In Section~\ref{sec:scaled_ei} we derive the variance
of the improvement quantifier and use it to define a new acquisition
function. It improves the literature ones by accounting for another
layer of uncertainty in the problem, namely the uncertainty in the
improvement random variable.  Sections~\ref{sec:benchmark-study}
and \ref{sec:benchmark-results} compare the newly introduced acquisition
function with state-of-the-art acquisition functions from the Bayesian
optimization literature on a large set of test functions for global
optimization. In Section \ref{sec:Comparative-Study} we confirm that
Bayesian optimization reduces the number of function evaluations required
to reach the global optimum to a certain tolerance level, compared
to standard global optimization algorithms. The proposed algorithm
is then used in Section \ref{sec:Pulmonary} to infer the parameters
of a partial differential equations (PDEs) model of the human pulmonary blood circulation system, 
with the ultimate goal to pave the way towards autonomous in silico diagnosis and prognosis.
Section~\ref{sec:conclusions} summarizes our results.

\section{Background\label{sec:background}}

Suppose that the task is to minimize globally a real-valued function
$f(\xb)$, called \emph{objective function,} over a compact domain
$\mathbb{X}\subset\bb R^{d}$ and that observing $f(\xb)$ is costly
due to the need to run long computer simulations or physical experiments.
The global minimum is denoted $\fglob=\min_{\xb\in\bb X}f(\xb)$,
which is attained at $\xglob$. Here we will focus on the minimization
problem as the conversion of a maximization problem into a minimization
one is trivial. Many global optimization algorithms have been proposed
in literature, e.g. genetic algorithms, multistart and simulated annealing
methods~\citep{Locatelli2013}, but these algorithms require many
function evaluations and hence are designed for functions that are
cheap to query. Bayesian optimization (BO), instead, is an algorithm
designed to optimize expensive-to-evaluate functions by keeping the
number of function evaluations as low as possible, hence saving computational
time. To do so, BO uses all of the information collected so far (function
values and corresponding locations) to internally maintain a model
of the objective function. This is used to learn about the location
of the minimum, and the model is continuously updated as new information
arrives. The objective function $f$ is approximated by a \emph{surrogate
model} or \emph{emulator}, which is usually given a Gaussian process
(GP) prior, see~\citet{Rasmussen2006}. The values of the objective
function are generally modelled according to the additive decomposition
$y_{i}=f(\xb_{i})+\e_{i}$, where $\e_{i}$ are i.i.d. $\N(0,\sigma^{2})$
errors and $f\sim\GP mk$ is the GP prior on the regression function.

\subsection{Gaussian Processes\label{subsec:gaussian_processes}}

A random process $\left\{ f(\xb),\,\xb\in\mathbb{X}\right\} $ is
said to be Gaussian if and only if every finite dimensional distribution
is a Gaussian random vector. Similarly to a multivariate Normal, parametrized
by a mean vector and a covariance matrix, a Gaussian process is completely
specified by a mean and a covariance function, denoted by $m(\boldsymbol{x})$
and $k(\boldsymbol{x},\boldsymbol{x}')$ respectively. They return
the mean $\E[f(\boldsymbol{x})]=m(\boldsymbol{x})$ and the covariance
$\cov{f(\boldsymbol{x})}{f(\boldsymbol{x}')}=k(\boldsymbol{x},\boldsymbol{x}')$
as function of the index only. A GP prior on the random function $f$
is denoted $f(\xb)\sim\GP{m(\xb)}{k(\xb,\xb')}$. In our studies we
used a constant mean function: $m(\xb)=c$, where $c$ represents
a constant to be estimated. The covariance functions considered are
the ARD Mat\'{e}rn 5/2 kernel:
\[
k_{5/2}(\xb,\xb')=\sigma_{f}^{2}\left(1+\sqrt{5}r+\frac{5}{3}r^{2}\right)\exp\left(-\sqrt{5}r\right),\qquad r=\sqrt{\sum_{i=1}^{d}\frac{(x_{i}-x_{i}')^{2}}{\ell_{i}^{2}}},
\]
and the ARD Squared Exponential kernel:
\[
k_{\mathrm{SE}}(\xb,\xb')=\sigma_{f}^{2}\exp\left\{ -\frac{1}{2}\sum_{i=1}^{d}\frac{(x_{i}-x_{i}')^{2}}{\ell_{i}^{2}}\right\} ,
\]
see \citet{Rasmussen2006} for more details, and the two kernels gave very similar experimental results (compare Sections \ref{sec:benchmark-results} and \ref{subsec:Matern-5/2-kernel}). 
The acronym ARD stands for automatic relevance determination,
and it refers to kernels that allow for a different lengthscale parameter
$\ell_{i}$ $(i=1,\dots,d)$ in each dimension. When an input dimensionality
has a large lengthscale, the function is effectively flat along that
direction, meaning that the input is not relevant. The model hyperparameters,
$\boldsymbol{\theta}=(c,\ell_{1},\dots,\ell_{d},\sigma_{f},\sigma)^{\T}$,
are estimated by maximum marginal likelihood. In the main part of
the paper we present results for the widely used ARD Squared Exponential
kernel, in order to allow for a fair comparison with the competing
methods, whose code is provided only for this family of covariance
functions. 
A GP with the ARD Squared Exponential kernel models infinitely differentiable functions, while the ARD Mat\'{e}rn 5/2 kernel gives rise to sample paths which are only twice differentiable.
Results for the ARD Mat\'{e}rn 5/2 kernel, recommended
by \citet{Snoek2012}, are shown in Appendix \ref{subsec:Matern-5/2-kernel}.
After having collected $n$ data points, $\D_{n}=\{(\xb_{1},y_{1}),\dots,(\xb_{n},y_{n})\}$,
the predictive distribution of $f$ is again a GP with mean $\fhat(\boldsymbol{x})$
and covariance $s(\boldsymbol{x},\xb')$: 
\begin{align}
f(\xb)\mid\D_{n} & \sim\GP{\fhat(\xb)}{s(\xb,\xb')}\label{eq:gp_predictive_distribution}\\
\fhat(\boldsymbol{x}) & =m(\xb)+\boldsymbol{k}(\boldsymbol{x})^{\T}[\boldsymbol{K}+\sigma^{2}\boldsymbol{I}]^{-1}(\boldsymbol{y}-\boldsymbol{m})\nonumber \\
s(\boldsymbol{x},\boldsymbol{x}') & =k(\boldsymbol{x},\boldsymbol{x}')-\boldsymbol{k}(\boldsymbol{x})^{\T}[\boldsymbol{K}+\sigma^{2}\boldsymbol{I}]^{-1}\boldsymbol{k}(\boldsymbol{x}'),\nonumber 
\end{align}
where $\boldsymbol{K}=[k(\boldsymbol{x}_{i},\boldsymbol{x}_{j})]_{i,j=1}^{n}$
is the $n$-by-$n$ covariance matrix at the training inputs, $\boldsymbol{k}(\boldsymbol{x})=[k(\boldsymbol{x}_{1},\boldsymbol{x}),\dots,k(\boldsymbol{x}_{n},\boldsymbol{x})]^{\T}$
is the column vector of size $n$-by-$1$ containing the covariances
between each of the training inputs and the test point $\xb$, while
$\boldsymbol{m}=[m(\xb_{1}),\dots,m(\xb_{n})]^{\T}$ represents the
mean at the training inputs. The posterior variance is readily obtained
as $s^{2}(\boldsymbol{x})=s(\boldsymbol{x},\boldsymbol{x})=\cov{f(\xb)}{f(\xb)}$.

\subsection{Bayesian Optimization\label{subsec:bayesian_optimization}}

The strength of BO lies in the following problem shift. Instead of
directly optimizing the expensive objective function $f$, the optimization
is performed on an inexpensive auxiliary function which uses the available
information in order to recommend the next query point $\xnext$,
hence it is referred to as \emph{acquisition function}. Optimization
algorithms propose a sequence $\xb_{n}$ of points that aim to converge
to a global optimum $\xglob$. In order to propose such a sequence,
BO algorithms start by evaluating the objective function $f$ at an
initial design: $\D_{\ninit}=\{(\xb_{i},y_{i})\}_{i=1}^{\ninit}$.
\citet{Jones1998} recommend to use a space filling Latin hypercube
design of $\ninit=10\times d$ points, with $d$ being the dimensionality
of the input space. Then iterate until the maximum number of function
evaluations $n_{\mathrm{max}}$ is reached: 1) obtain the predictive
distribution of $f$ given $\D_{n}$, 2) use the distribution of $f$
given $\D_{n}$ to compute the auxiliary function $a_{n}(\xb)$, 3)
solve the auxiliary optimization problem $\xb_{\next}=\argmax a_{n}(\boldsymbol{x})$,
4) query $f$ at the recommended point $\xnext$ and update the training
data: $\D_{n+1}=\D_{n}\cup\{\xnext,\fnext\}$. For a pseudocode-style
algorithm, see Algorithm~\ref{alg:bo}.

\begin{algorithm}[tb]
   \caption{Bayesian optimization.\label{alg:bo}}
\begin{algorithmic}[1]
   \STATE \textbf{Inputs:} \\
   Initial design: $\D_{\ninit} = \{ (\xb_i, y_i) \}_{i=1}^{\ninit}$ \\
   Budget of $ \nmax $ function evaluations
   \item[]
   \FOR{$n = \ninit$ \textbf{to} $\nmax - 1$}
   \STATE Update the GP: $f(\xb) \mid \D_{n} \sim \GP{\fhat(\xb)}{s(\xb,\xb')}$
   \STATE Compute the acquisition function: $ a_n(\xb) $
   \STATE Solve the auxiliary optimization problem: $\xnext = \argmax_{\xb \in \bb{X}} a_n(\xb)$
   \STATE Query $f$ at $\xnext$ to obtain $\fnext$
   \STATE Augment data: $\D_{n + 1} = \D_{n} \cup \{ \xnext, \fnext\}$
   \ENDFOR
   \item[]
   \STATE \textbf{Return:} \\
   Estimated minimum: $f_\mathrm{min} = \min(y_1, \dots, y_{\nmax})$ \\
   Estimated point of minimum: $\boldsymbol{x}_\mathrm{min} = \argmin(y_1, \dots, y_{\nmax})$
\end{algorithmic}
\end{algorithm}

Different BO algorithms vary in the choice of the acquisition function.
These can be grouped into three main categories: \emph{optimistic},
\emph{improvement-based} and \emph{information-based}~\citep{Shahriari2016}.
All acquisition functions try to balance to a different extent the
concepts of \emph{exploitation} and \emph{exploration}. The former
indicates evaluating where the emulator predicts a low function value,
while the latter means reducing our uncertainty about the model of
$f$ by evaluating at points of high predictive variance.

\emph{Optimistic policies} (class 1) handle exploration and exploitation
by being optimistic in the face of uncertainty, in the sense of considering
the best case scenario for a given probability value. The approach
of \citet{Cox1997} was to consider a \emph{statistical lower bound}
on the minimum, $\mathrm{LCB}(\xb)=-\{\fhat(\xb)-\kappa s(\xb)\}$,
where the minus sign in front is needed as the acquisition function
is maximized. This acquisition function is known as the lower confidence
bound (LCB) policy. 
Here, $\kappa$ is a parameter managing the trade-off between \emph{exploitation} and \emph{exploration}. 
When $\kappa=0$, the focus is on pure exploitation, i.e. evaluating where the GP model predicts low function values. 
On the contrary, a high value of $\kappa$ emphasizes exploration by inflating the model uncertainty, i.e. recommending to evaluate at points of high predictive uncertainty. 
For this acquisition function there are strong
theoretical results on achieving the optimal regret derived by \citet{Srinivas2012}.

The next group (class 2) of acquisition functions are \emph{improvement-based}.
Define the current best function value at iteration $n$ to be $\fmin=\min(y_{1},\dots,y_{n})$\footnote{If the function values are corrupted by noise, $\fmin=\min\fhat(\xb)$.},
and recall that $f(\xb)\mid\D_{n}\sim\N(\fhat(\xb),s^{2}(\xb))$ from
the marginalization property of GPs. By standardization, $z(\xb)=\{f(\xb)-\fhat(\xb)\}/s(\xb)$
has a standard normal distribution. This class of functions is based
on the random variable \emph{Improvement:} 
\begin{equation}
I(\xb)=\max\{\fmin-f(\xb),0\}.\label{eq:improvement}
\end{equation}
Intuitively, $I(\xb)$ assigns a reward of $\fmin-f(\xb)$ if $f(\xb)<\fmin$,
and zero otherwise. \citet{Kushner1964} proposed to select the point
that has the highest probability of improving upon the current best
function value $\fmin$. This effectively corresponds to maximizing
the probability of the event $\{I(\xb)>0\}$ or, equivalently, of
$\{f(\xb)<\fmin\}$. Define $u=\{\fmin-\fhat(\xb)\}/s(\xb)$. The
\emph{Probability of Improvement} (PI) acquisition function is:
\begin{align}
\mathrm{PI}(\xb) & =\bb P\{I(\xb)>0\}\nonumber \\
 & =\E1_{\{f(\xb)<\fmin\}}\nonumber \\
 & =\bb P\{f(\xb)<\fmin\}\label{eq:probability_of_improvement}\\
 & =\bb P\{z(\xb)<u\}\nonumber \\
 & =\Phi(u).\nonumber 
\end{align}
For a detailed derivation see Section~\ref{subsec:derivation_pi}.
In the following, $\phi(x\mid\mu,\sigma^{2})$ and $\Phi(x\mid\mu,\sigma^{2})$
denote the probability density function (pdf) and cumulative distribution function (cdf) of a $\N(\mu,\sigma^{2})$ random variable.
For brevity, when $\mu=0$ and $\sigma^{2}=1$ we will simply write
$\phi(x)$ and $\Phi(x)$. The PI acquisition function corresponds
to the expectation of the utility $\mathfrak{u}(\xb)=1_{\{f(\xb)<\fmin\}}$,
which is $\mathfrak{u}(\xb)=1$ when $f(\xb)<\fmin$ and $0$ otherwise.
In other words, the utility assigns a reward of $1$ when we have
an improvement, irrespective of the magnitude of this improvement,
and $0$ otherwise. It might seem naive to assign a reward always
equal to $1$ every time we improve on $\fmin$, irrespectively of
the value. An acquisition function that accounts for the magnitude
of the improvement is obtained by averaging over the utility $\mathfrak{u}(\xb)=\fmin-f(\xb)$
when $f(\xb)<\fmin$ and $0$ otherwise, hence $\mathfrak{u}(\xb)=I(\xb)$.
The \emph{Expected Improvement} (EI) acquisition function \citep{Jones1998}
corresponds to the expectation of the random variable $I(\xb)$ and
is equal to:
\begin{alignat}{1}
\mathrm{EI}(\xb) & =\E\{I(\xb)\}\nonumber \\
 & =\{\fmin-\fhat(\xb)\}\Phi(u)+s(\xb)\phi(u),\label{eq:expected_improvement}
\end{alignat}
 see Section~\ref{subsec:derivation_ei} for a derivation. This policy
recommends to query at the point where we expect the highest improvement
score over the current best function value. The EI is made up of two
terms. The first term is increased by decreasing the predictive mean
$\fhat(\xb)$, the second term is increased by increasing the predictive
uncertainty $s(\xb)$. This shows how EI \emph{automatically} balances \emph{exploitation}
and \emph{exploration}.

Recent interest has focused on \emph{information-based} acquisition
functions (class 3). Here, the core idea is to query at points that
can help us learn more about the location of the unknown minimum rather
than points where we expect to obtain low function values. The main
representatives of this class are Entropy Search (ES) \citep{Hennig2012},
Predictive Entropy Search (PES) \citep{Hernandez-Lobato2014} and,
more recently, Max-Value Entropy Search (MES) \citep{Wang2017}. Both
ES and PES focus on the distribution of the argmin, $p(\xglob\mid\D_{n})$,
which is induced by the GP prior on $f$. These two policies recommend
to query at the point $\xnext$ leading to the largest reduction in
uncertainty about the distribution $p(\xglob\mid\D_{n})$. This can
be expressed as selecting the point $\{\xb,y\}$ conveying the most
information about $\xglob$ in terms of the mutual information $\bb I(\{\xb,y\},\xglob\mid\D_{n})$.
The Entropy Search acquisition function is $\mathrm{ES}(\xb)=H[\xglob\mid\D_{n}]-\E\{H[\xglob\mid\D_{n},\xb,y]\}$,
where $H$ is the entropy and the expectation is taken with respect
to the density $p(y\mid\D_{n},\xb)$. PES, instead, uses the symmetry
of mutual information in order to obtain the equivalent formulation:
$\mathrm{PES}(\xb)=H[y\mid\D_{n},\xb]-\E\{H[y\mid\D_{n},\xb,\xglob]\}$,
where the expectation is with respect to $p(\xglob\mid\D_{n})$. This
distribution is analytically intractable, and so is its entropy, hence calling for approximations based on a discretization of the input space, which incurs a loss of accuracy, and Monte Carlo sampling, which is computationally expensive. 
Furthermore, the point at which
the global minimum is attained might not be unique. Instead of measuring
the information about $\xglob$, which lies in a multidimensional
space $\bb X$, \citet{Wang2017} propose to focus on the simpler
gain in information between $y$ and the minimum value $\fglob,$
which lies in a one-dimensional space. The acquisition function hence
becomes $\mathrm{MES}(\xb)=\bb I(\{\xb,y\},\fglob\mid\D_{n})=H[y\mid\D_{n},\xb]-\E\{H[y\mid\D_{n},\xb,\fglob]\}$,
with expectation with respect to $p(\fglob\mid\D_{n})$. The expectation
is approximated with Monte Carlo estimation by sampling a set of function
minima. In summary the methods in this class involve 1) hyperparameters
sampling for marginalization and 2) sampling global optima for entropy
estimation. Step 2 substantially increases the computational cost
of information-based acquisition functions, especially in the case
of ES and PES, which sample in a multidimensional space.

\section{On a new Improvement-Based Acquisition Function\label{sec:scaled_ei}}

In the previous section it was shown that the widely used \emph{Expected
Improvement} (EI) acquisition function automatically balances \emph{exploitation}
and \emph{exploration}. What it does not account for, however, is
the uncertainty in the improvement value $I(\xb)$. This might not
be ``orthogonal'' to the uncertainty in the model of $f$, but it
nevertheless represents an important source of information about our
belief in the quality of a candidate point $\xb$. Ideally, to avoid
unnecessary expensive function evaluations, we hope to evaluate at
points where, on average, the improvement is expected to be high,
with high confidence. This is to avoid expensive queries at points
where the improvement is high, but the variability of this value is
also high, effectively meaning that the improvement score $I(\xb)$
could be low, and thus we would evaluate at a sub-optimal point $\xb$.
 In order to reach such a goal, we derive the variance of the improvement
quantifier $I(\xb)$: 
\begin{equation}
\V[I(\xb)]=s^{2}(\xb)\{(u^{2}+1)\Phi(u)+u\phi(u)\}-\{\mathrm{EI}(\xb)\}^{2},\label{eq:variance_of_improvement}
\end{equation}
where, again, $u=\{\fmin-\fhat(\xb)\}/s(\xb)$.
\begin{proof}
The proof uses properties (\ref{eq:phi_d}) and (\ref{eq:phi_dd})
of Gaussian pdfs from Appendix~\ref{subsec:derivatives_phi}:

\begin{alignat*}{1}
\V[I(\xb)] & =\E[I^{2}(\xb)]-\{\E[I(\xb)]\}^{2}\\
 & =\E[\max\{\fmin-f(\xb),0\}^{2}]-\{\mathrm{EI}(\xb)\}^{2}\\
 & =\int_{-\infty}^{\fmin}\{\fmin-y\}^{2}\phi(y\mid\fhat(\xb),s^{2}(\xb))\d y-\{\mathrm{EI}(\xb)\}^{2}\\
 & =\int_{-\infty}^{u}\{\fmin-\fhat(\xb)-s(\xb)z\}^{2}\phi(z)\d z-\{\mathrm{EI}(\xb)\}^{2}\\
 & =\int_{-\infty}^{u}\{[\fmin-\fhat(\xb)]^{2}+z^{2}s^{2}(\xb)\\
 & \qquad-2zs(\xb)[\fmin-\fhat(\xb)]\}\phi(z)\d z-\{\mathrm{EI}(\xb)\}^{2}\\
 & =\{\fmin-\fhat(\xb)\}^{2}\int_{-\infty}^{u}\phi(z)\d z\\
 & \qquad+s^{2}(\xb)\int_{-\infty}^{u}z^{2}\phi(z)\d z\\
 & \qquad-2s(\xb)\{\fmin-\fhat(\xb)\}\int_{-\infty}^{u}z\phi(z)\d z-\{\mathrm{EI}(\xb)\}^{2}  \\
 & =\{\fmin-\fhat(\xb)\}^{2}\Phi(u)+2s(\xb)\{\fmin-\fhat(\xb)\}\phi(u)\\
 & \qquad+s^{2}(\xb)\int_{-\infty}^{u}(z^{2}-1)\phi(z)\d z\\
 & \qquad+s^{2}(\xb)\int_{-\infty}^{u}\phi(z)\d z-\{\mathrm{EI}(\xb)\}^{2}\\
 & =\{\fmin-\fhat(\xb)\}^{2}\Phi(u)+2s(\xb)\{\fmin-\fhat(\xb)\}\phi(u)\\
 & \qquad-s^{2}(\xb)u\phi(u)+s^{2}(\xb)\Phi(u)-\{\mathrm{EI}(\xb)\}^{2}.
\end{alignat*}
From the definition of $u=\{\fmin-\fhat(\xb)\}/s(\xb)$, we obtain:
\begin{alignat*}{1}
\V[I(\xb)] & =[\{\fmin-\fhat(\xb)\}^{2}+s^{2}(\xb)]\Phi(u)\\
 & \qquad+s(\xb)\{\fmin-\fhat(\xb)\}\phi(u)-\{\mathrm{EI}(\xb)\}^{2}\\
 & =s^{2}(\xb)\{(u^{2}+1)\Phi(u)+u\phi(u)\}-\{\mathrm{EI}(\xb)\}^{2}.
\end{alignat*}
\end{proof}

We now define a new acquisition function that we call the \emph{Scaled
Expected Improvement} (ScaledEI): 
\begin{equation}
\mathrm{ScaledEI}(\xb)=\E[I(\xb)]/\{\V[I(\xb)]\}^{1/2}.\label{eq:scaled_ei}
\end{equation}
Selecting the next query point by maximizing this acquisition function
corresponds to selecting query points where the improvement score
is expected to be high with high confidence.

\section{Benchmark Study\label{sec:benchmark-study}}

\begin{figure}[tb]
\begin{centering}
\includegraphics[width=\textwidth]{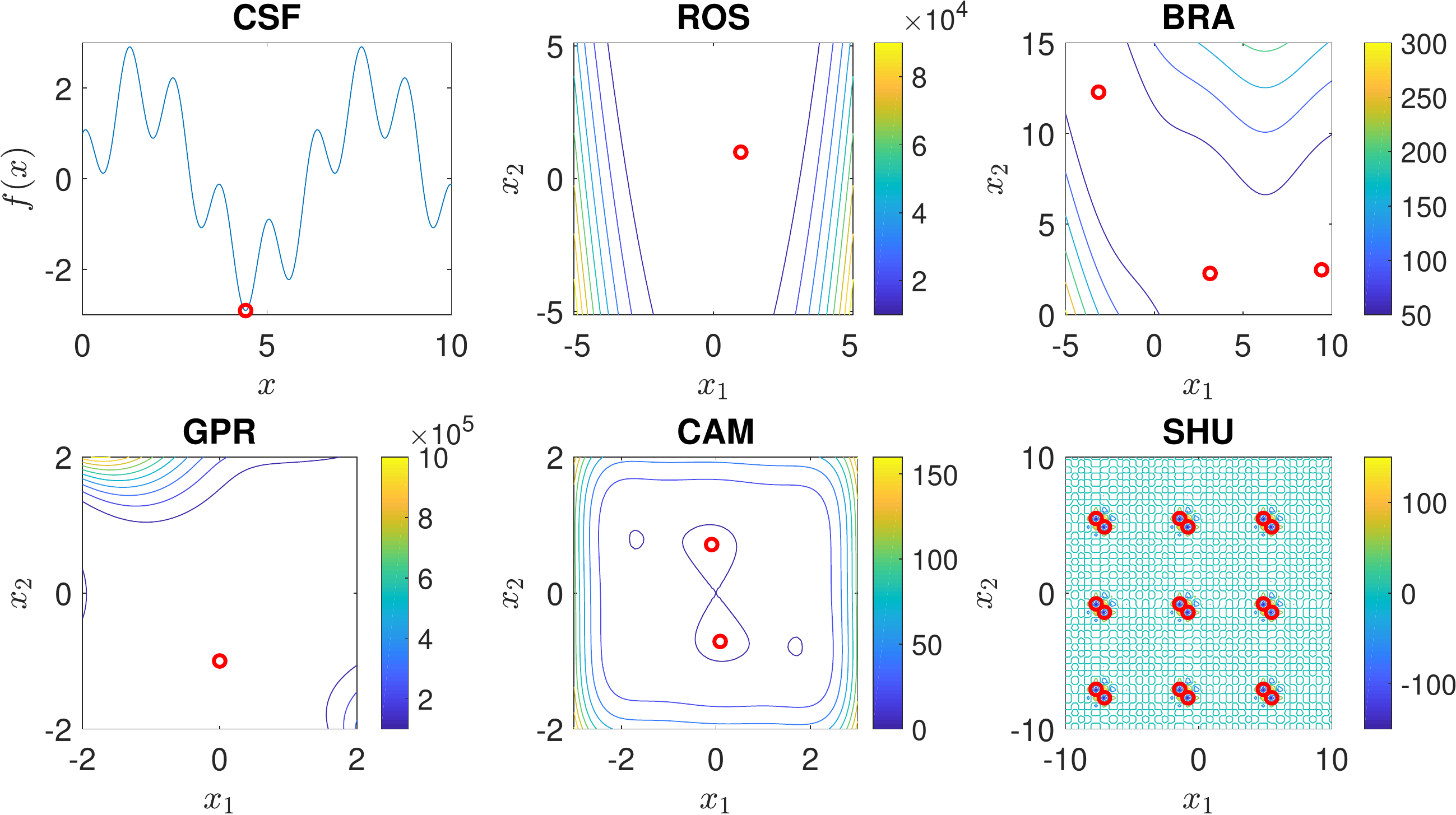}
\par\end{centering}
\caption{\textbf{Plot of the 1D function CSF and contours of the 2D test functions.}
The red circles represent the global optima.\label{fig:test_functions}}
\end{figure}

We test the performance of the acquisition function introduced in
(\ref{eq:scaled_ei}) and the ones from the BO literature on an extensive
test set of objective functions taken from the global optimization
literature, summarized in Table~\ref{tab:test_functions}. This comprehensive
test set includes benchmark functions found in leading global optimization
articles such as \citet{Jones1993} and \citet{Huyer1999}, with the
addition of the 1D Cosine Sine (CSF) test function, which we defined as $f(x)=\cos(5x)+2\sin(x)$.
These optimization problems are challenging due to the presence of
multiple local minima, the sharp variation in the $y$-axis, and symmetries
with the presence of multiple points at which the global minimum is
attained. Figure~\ref{fig:test_functions} shows a plot of the 1D
test function CSF and the contours of the 2D objective functions along
with the global optima.

\begin{table}[tb]
\centering{}\caption{\textbf{Key characteristics of the test functions.}\label{tab:test_functions}}
\begin{tabular}{lcccc}
\hline 
 &  & Number of  & Number of  & Number of\tabularnewline
Test function  & Abbreviation  & dimensions  & local minima  & global minima\tabularnewline
\hline 
Cosine Sine  & CSF  & 1  & 8  & 1\tabularnewline
Rosenbrock  & ROS  & 2  & 1  & 1\tabularnewline
Branin RCOS  & BRA  & 2  & 3  & 3\tabularnewline
Goldstein and Price  & GPR  & 2  & 4  & 1\tabularnewline
Six-Hump Camel  & CAM  & 2  & 6  & 2\tabularnewline
Two-Dimensional Shubert  & SHU  & 2  & 760  & 18\tabularnewline
Hartman 3  & HM3  & 3  & 4  & 1\tabularnewline
Shekel 5  & SH5  & 4  & 5  & 1\tabularnewline
Shekel 7  & SH7  & 4  & 7  & 1\tabularnewline
Shekel 10  & SH10  & 4  & 10  & 1\tabularnewline
Hartman 6  & HM6  & 6  & 4  & 1\tabularnewline
Rastrigin  & RAS  & 10  & 11\textsuperscript{10} & 1\tabularnewline
\hline 
\end{tabular}
\end{table}

Let $\fglob$ denote the globally optimal function value known from
the literature and denote by $\fmin$ the best function value at iteration
$n$. To check for convergence to the global minimum $\fglob$ we
report the log\textsubscript{10} distance, defined as follows: 

\[
\text{log}_{10}\text{ distance}=\log_{10}|\fmin-\fglob|,
\]
where the dependency of the distance on $n$ comes through $\fmin$.

The established acquisition functions that we compared with the proposed
new acquisition function, ScaledEI, are LCB, representing the optimistic
policies (class 1); PI and EI from the improvement-based ones (class
2) and MES representing the information-theoretic measures (class
3), which has been shown to outperform ES and PES; see \citet{Wang2017}.
We also include as benchmarks two naive approaches: $\mathrm{RND}(\xb)$
and $\mathrm{MN}(\xb)$. The first corresponds to random search \citep{Bergstra2011,Bergstra2012},
which proposes a point from a uniform distribution within the bounded
domain $\bb X$. The second corresponds to iteratively maximizing
the negative GP predictive mean, $\mathrm{MN}(\xb)=-\fhat(\xb)$,
and is the extreme case of focusing only on \emph{exploitation} while
ignoring uncertainty. The GP model uses an ARD Squared Exponential
kernel and a constant mean function. The model hyperparameters are
estimated at each iteration by maximum marginal likelihood using the
Quasi Newton method. The maximization of the acquisition function
is performed by evaluating it at $10^{4}$ uniform points in the input
domain. The inputs are then ranked by their acquisition function values,
and the 10 points having the highest score are found. A Nelder-Mead
local solver is run starting from each of these 10 points, until each
solver reaches a relative tolerance on the function value of $10^{-3}$.
The next evaluation point $\boldsymbol{x}_\mathrm{next}$ is set to the best argmin returned by the 10 local solvers, while the remaining points are discarded.

For all experiments we set a priori the maximum budget of function
evaluations to be $\nmax=1000$, and an upper bound on the computational
runtime for the computer cluster\footnote{The computer cluster used in this work includes eight CentOS~7 machines
with 24 cores and 32GB RAM each.} of 2 weeks. The maximum budget is usually set by the analyst, considering
the unit price for a single evaluation. It could be an actual monetary
price, if the experiment involves materials and trained staff, or
computational resources. The code for the acquisition function MES
has been cloned as of July 2017 from \citet{Wang2017} first author's
GitHub repository\footnote{\url{https://github.com/zi-w/Max-value-Entropy-Search}}.
Since July 2017 the authors have not made significant changes to the
core functionality apart from some documentation updates. For the
experiments, the settings have been kept the same as in \citet{Wang2017}
in terms of GP mean and kernel choice. For optimal performance and
a fair comparison with the other methods, the GP hyperparameters were
updated at every iteration instead of their default choice of every
10 iterations. Most importantly, between the two versions of MES presented
by the authors (MES-R and MES-G), we chose the version that in their
paper was shown to perform best, namely the MES-G acquisition function
with $\fglob$ sampled from the approximate Gumbel distribution. The
choice was also motivated by the statement in \citet{Wang2017} that
MES-R is better for problems with only a few local optima, while MES-G
works better in highly multimodal problems as more exploration is
needed. As the set of test functions used is characterized by high
multimodality and the presence of multiple points at which the global
minimum is attained, we present results for the MES-G policy, which
will be simply denoted as MES. The number of $\fglob$ sampled was
set to 100 as in the experiments of \citet{Wang2017}. For the LCB
acquisition function, representing the class of optimistic policies,
we set $\kappa=2$ as commonly used, see
for example \citet{Turner2012}. We ran Bayesian optimization using
each of the acquisition functions on every benchmark function, and we repeated each optimization five times using different random number seeds for the initial design.

\section{Benchmark Results\label{sec:benchmark-results}}

\begin{figure}[tb]
\begin{centering}
\includegraphics[trim={1cm 1cm 1cm 1cm},clip,width=\textwidth]{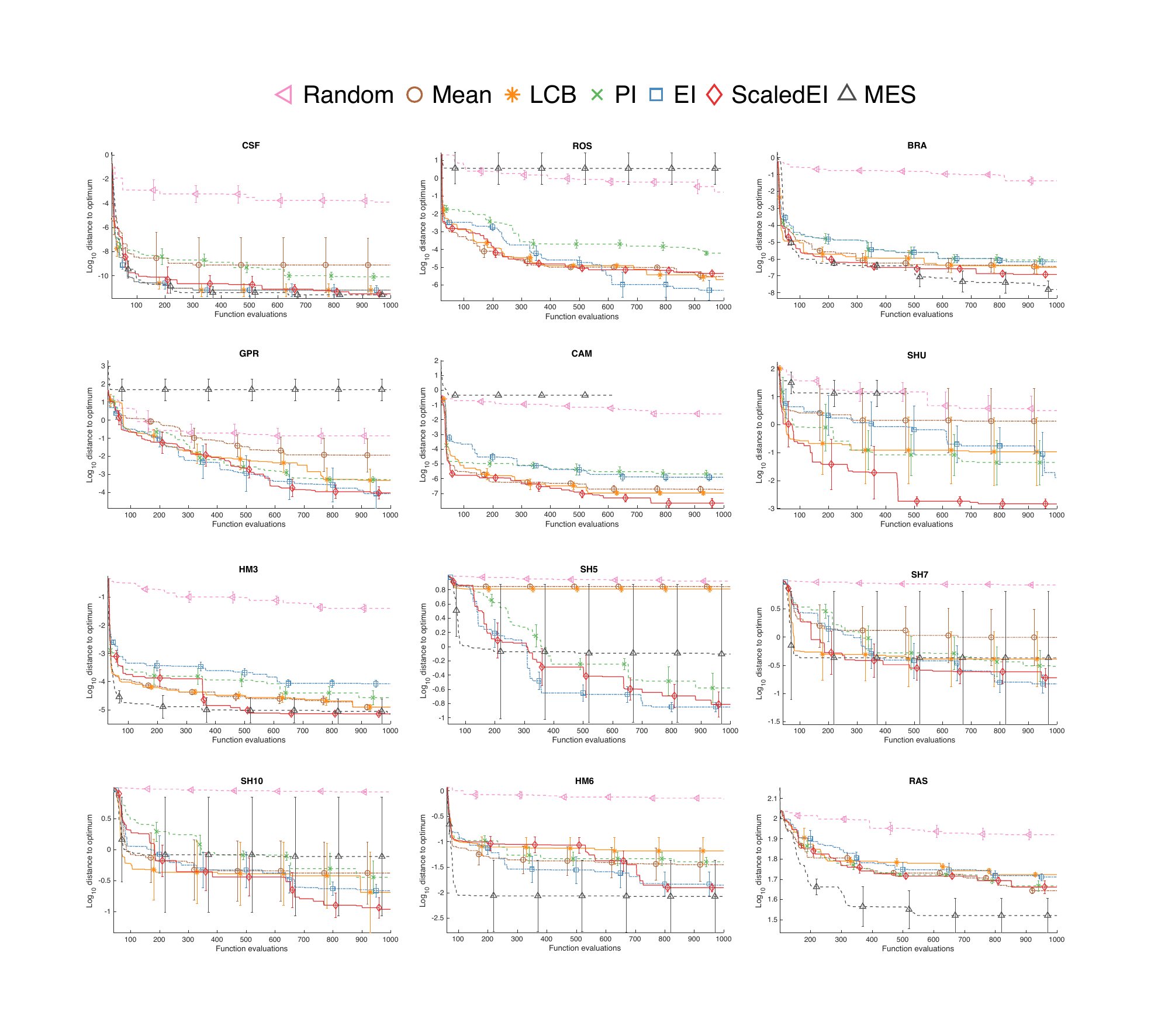}
\par\end{centering}
\caption{\textbf{Comparison of the log}\protect\textsubscript{\textbf{10}}\textbf{
distances (in function space) to the global optimum.} Each panel represents
a given benchmark function. The traces show the average distance over
the five design instantiations for the whole spectrum of iterations,
while the error bars show plus or minus one standard error of the
mean. \label{fig:se_logdistances}}
\end{figure}

This section empirically shows that the proposed acquisition function
performs as well as or better than the state-of-the-art methods reviewed
in Section~\ref{subsec:bayesian_optimization}, using an extensive
set of benchmark problems from the global optimization literature
\citep{Jones1993,Huyer1999}. Figure~\ref{fig:se_logdistances}
shows the log\textsubscript{10} distance in function space to the
global optimum as a function of the objective function evaluations
$n$. Each trace represents the average log\textsubscript{10} distance
of a given algorithm over the 5 different initial design instantiations, and
the error bars show the standard error of the mean for the whole spectrum
of iteration numbers. For two of the problems, CAM and SHU, the Max-Value
Entropy Search method did not reach the maximum budget $n_{\max}$
in the allowed computational time (2 weeks). In these cases the log\textsubscript{10}
distance in function space is shown up to the latest iteration.

\begin{table}[tbh]
\begin{centering}
\caption{\textbf{Statistical test for the significance in the mean difference
of the final log}\protect\textsubscript{\textbf{10}}\textbf{ distances.
}The ScaledEI acquisition function was tested against all remaining
acquisition functions using a paired t-test with significance level
0.05. Codes: 0 indicates a non significant difference and 1 (-1) indicates
that ScaledEI performed better (worse), i.e. it has a significantly
lower (higher) average distance.\label{tab:t_test_1000_se}}
\par\end{centering}
\centering{}%
\begin{tabular}{lcccccc}
\hline 
 & \multicolumn{6}{c}{ScaledEI vs}\tabularnewline
Test function  & RND  & MN  & LCB  & PI  & EI  & MES\tabularnewline
\hline 
    CSF        & 1 & 0 & 0 & 1 & 0 & 0\tabularnewline
   ROS        & 1 & 0 & 0 & 1 & 0 & 1\tabularnewline
   BRA        & 1 & 0 & 0 & 0 & 1 & 0\tabularnewline
    GPR        & 1 & 0 & 0 & 0 & 0 & 1\tabularnewline
   CAM        & 1 & 0 & 0 & 1 & 1 & 1\tabularnewline
   SHU        & 1 & 0 & 0 & 0 & 0 & 1\tabularnewline
HM3 & 1 & 0 & 0 & 0 & 1 & 0\tabularnewline
SH5 & 1 & 1 & 1 & 0 & 0 & 0\tabularnewline
SH7 & 1 & 0 & 0 & 0 & 0 & 0\tabularnewline
SH10 & 1 & 0 & 0 & 0 & 0 & 0\tabularnewline
HM6 & 1 & 0 & 0 & 1 & 0 & 0\tabularnewline
RAS       & 1 & 0 & 0 & 0 & 0 & 0\tabularnewline
\hline 
Same       & 0\% & 92\% & 92\% & 67\% & 75\% & 67\%\tabularnewline
Better     & 100\% & 8\% & 8\% & 33\% & 25\% & 33\%\tabularnewline
Worse      & 0\% & 0\% & 0\% & 0\% & 0\% & 0\%\tabularnewline
\hline 
\end{tabular}
\end{table}

The results in Figure~\ref{fig:se_logdistances} show
that the RND policy is consistently outperformed by the improvement-based
acquisition functions. In the 1D scenario (CSF) RND is the worst method,
followed by MN, which has huge variation in the results. The best
methods include improvement-based policies and MES, but there does
not seem to be a significant difference between them. For most of
the 2D functions (ROS, GPR, CAM, SHU) it appears that improvement-based
policies outperform the information-theoretic one. In three of these
functions the proposed ScaledEI outperforms all others. In the BRA
function, MES seems to be the best, but by a small margin compared
to the evident gap in the other 2D scenarios. In the 3D problem (HM3)
EI is outperformed by the other methods, except for RND. However,
the proposed method, ScaledEI, is the best, emphasizing the power
of the proposed adjustment. Even in the 4D, 6D and 10D test functions
the ScaledEI policy appears to be one of the most competitive methods,
while its information-based competitor MES suffers from highly variable
results.

In order to facilitate the interpretation of the results, we test
in Table~\ref{tab:t_test_1000_se} the significance of the difference
in means of the final log\textsubscript{10} distances\footnote{These are the distances at the last iteration, where either the pre-defined
budget or the maximum CPU time was exceeded.} for ScaledEI vs each of the remaining acquisition functions, using
a paired t-test with significance level 0.05. We remark that the choice
of performing a t-test on the log distances at the final iteration
is only for summary purposes, and Figure \ref{fig:se_logdistances}
shows the full spectrum of performance scores for all function evaluations
ranging from $n=100$ to $1000$. In most cases, the log distance
curves are fairly flat between $300$ and $1000$, so the table presented
here is representative of the majority of the choices of $n$. In general we would not get the true optimum at a high degree of accuracy
with only $200$ function evaluations. 
Nevertheless, we have carried out the statistical hypothesis tests for $n=600$ and $n=200$ as well, and the results can be found in Tables \ref{tab:t_test_600_se} and \ref{tab:t_test_200_se}. 
A score of 0 indicates that the null hypothesis of equal average log\textsubscript{10} distance
is \emph{not} rejected. Both 1 and -1 indicate the rejection of the
null hypothesis, where a score of 1 indicates that the proposed method,
ScaledEI, achieves a significant improvement, while a score of -1
shows that ScaledEI is significantly worse.

The proposed acquisition function ScaledEI consistently outperforms
the naive RND policy. Compared with the established acquisition functions,
ScaledEI nearly always achieves equal (67-92\%) or significantly better
(8-33\%) performance, without being significantly outperformed by
the competitors. One third of the times, ScaledEI is significantly
better than PI and the information-theoretic competitor MES. This
is corroborated in Figure~\ref{fig:se_logdistances},
where for the benchmark functions ROS, GPR, CAM, SHU the information-theoretic
policy does not come close to the minimum. Then follow EI and LCB,
which are outperformed 25\% and 8\% of the times, and MN, which is,
surprisingly, outperformed only 8\% of the times. This finding provides
reassurance for conservative BO strategies, as in~\citet{Wang2013}.
However, the summary table based on hypothesis tests incurs a loss
of information. From the t-test it is not evident that the MN policy
suffers from huge variations in the results (see Figure~\ref{fig:se_logdistances}).
The fact that for some random number seeds MN performs well is due
to the initial design generating a point near the global minimum by
chance. Then, by emphasizing exploitation only, this point will be
fine-tuned to the global optimum. ScaledEI never appears to be significantly
worse than its competitors, and in particular the widely applied EI method.
Our results suggest that the proposed acquisition function, ScaledEI,
which combines high expected improvement with high confidence in the
improvement being high, performs as well as or better than state-of-the-art
acquisition functions. This makes ScaledEI a good default choice for
standard BO applications.

\begin{table}[tb]
\begin{centering}
\caption{\textbf{Statistical test for the significance in the mean difference
of the log}\protect\textsubscript{\textbf{10}}\textbf{ distances
at $\boldsymbol{n=600}$. }The ScaledEI acquisition function was tested
against all remaining acquisition functions using a paired t-test
with significance level 0.05. Codes: 0 indicates a non significant
difference and 1 (-1) indicates that ScaledEI performed better (worse),
i.e. it has a significantly lower (higher) average distance.\label{tab:t_test_600_se}}
\par\end{centering}
\centering{}%
\begin{tabular}{lcccccc}
\hline 
 & \multicolumn{6}{c}{ScaledEI vs}\tabularnewline
Test function  & RND  & MN  & LCB  & PI  & EI  & MES\tabularnewline
\hline 
 CSF  & 1 & 0 & 0 & 1 & 0 & 0\tabularnewline
 ROS  & 1 & 0 & 0 & 1 & 0 & 1\tabularnewline
 BRA  & 1 & 0 & 0 & 1 & 1 & 0\tabularnewline
 GPR  & 1 & 0 & 0 & 0 & 0 & 1\tabularnewline
 CAM  & 1 & 0 & 0 & 1 & 1 & 1\tabularnewline
 SHU  & 1 & 0 & 0 & 0 & 1 & 1\tabularnewline
HM3 & 1 & 0 & 0 & 1 & 1 & 0\tabularnewline
SH5 & 1 & 1 & 1 & 0 & 0 & 0\tabularnewline
SH7 & 1 & 0 & 0 & 0 & 0 & 0\tabularnewline
SH10 & 1 & 0 & 0 & 0 & 0 & 0\tabularnewline
HM6 & 1 & 0 & 0 & 0 & 0 & 0\tabularnewline
 RAS  & 1 & 0 & 1 & 0 & 0 & 0\tabularnewline
\hline 
 Same  & 0\% & 92\% & 83\% & 58\% & 67\% & 67\%\tabularnewline
 Better  & 100\% & 8\% & 17\% & 42\% & 33\% & 33\%\tabularnewline
 Worse  & 0\% & 0\% & 0\% & 0\% & 0\% & 0\%\tabularnewline
\hline 
\end{tabular}
\end{table}

As already mentioned above, we would not expect the algorithms to fine-tune the returned optimum in only $n=200$ steps. However, for representational completeness we have carried out the statistical hypothesis tests for $n=600$ and $n=200$ nevertheless. These are shown in Tables \ref{tab:t_test_600_se} and \ref{tab:t_test_200_se} respectively. 

Table \ref{tab:t_test_600_se} shows that at 600 iterations the results are consistent with Table \ref{tab:t_test_1000_se}, which uses the full budget of function evaluations. ScaledEI is always significantly better than RND, but only 8\% of the times better than MN. Again, this is due to the random generation of an initial design point near a global minimum by chance, which is fine-tuned by focusing on exploitation only. However, this approach carries substantial variations in the results, and the success of the MN method depends entirely on the initial design choice, making it a non-optimal policy. For the remaining methods, ScaledEI is significantly better than each competitor 17-42\% of the times, and in the remaining cases the methods are not significantly different. 
One third of the times it is better than the widely used EI method and the information-theoretic MES. Furthermore, ScaledEI is never significantly outperformed by its competitors, including MES. 
We finally report in Table \ref{tab:t_test_200_se} the same kind of test but stopping at $n=200$ iterations only. In this scenario, ScaledEI performed significantly better than PI and EI, 17\% and 42\% of the times respectively, while in the remaining cases the two were not significantly different. Comparing
ScaledEI and the information theoretic strategy (MES), 50\% of the
time the two are not significantly different, but 33\% ScaledEI is
performing significantly better, and it is outperformed by MES in
two benchmark functions only: HM3 and RAS. 

As remarked in Section \ref{sec:background},
different BO algorithms vary in the choice of the acquisition function.
The proposed one, ScaledEI, was tested against literature methods
on a set of 12 benchmark functions having different functional characteristics.
According to the no-free-lunch theorem, we do not expect to see one
method consistently outperforming all the remaining algorithms on
all benchmark functions and for any arbitrary choice of function evaluations $n$. 
However, our results, shown in Figure \ref{fig:se_logdistances} and Tables \ref{tab:t_test_1000_se} to \ref{tab:t_test_200_se}, suggest that ScaledEI tends to perform as well as or better than the alternative methods, and hence constitutes a powerful default choice for Bayesian optimization.
 
\begin{table}[tb]
\begin{centering}
\caption{\textbf{Statistical test for the significance in the mean difference
of the log}\protect\textsubscript{\textbf{10}}\textbf{ distances
at $\boldsymbol{n=200}$. }The ScaledEI acquisition function was tested
against all remaining acquisition functions using a paired t-test
with significance level 0.05. Codes: 0 indicates a non significant
difference and 1 (-1) indicates that ScaledEI performed better (worse),
i.e. it has a significantly lower (higher) average distance.\label{tab:t_test_200_se}}
\par\end{centering}
\centering{}%
\begin{tabular}{lcccccc}
\hline 
 & \multicolumn{6}{c}{ScaledEI vs}\tabularnewline
Test function  & RND  & MN  & LCB  & PI  & EI  & MES\tabularnewline
\hline 
 CSF  & 1 & 0 & 0 & 0 & 0 & 0\tabularnewline
 ROS  & 1 & 0 & 0 & 1 & 1 & 1\tabularnewline
 BRA  & 1 & 0 & 0 & 1 & 1 & 0\tabularnewline
 GPR  & 0 & 0 & 0 & 0 & 0 & 1\tabularnewline
 CAM  & 1 & 0 & 0 & 0 & 1 & 1\tabularnewline
 SHU  & 0 & 0 & 0 & 0 & 1 & 1\tabularnewline
HM3 & 1 & 0 & 0 & 0 & 1 & -1\tabularnewline
SH5 & 1 & 0 & 0 & 0 & 0 & 0\tabularnewline
SH7 & 1 & 0 & 0 & 0 & 0 & 0\tabularnewline
SH10 & 1 & 0 & 0 & 0 & 0 & 0\tabularnewline
HM6 & 1 & 0 & 0 & 0 & 0 & 0\tabularnewline
 RAS  & 1 & 0 & 0 & 0 & 0 & -1\tabularnewline
\hline 
 Same  & 17\% & 100\% & 100\% & 83\% & 58\% & 50\%\tabularnewline
 Better  & 83\% & 0\% & 0\% & 17\% & 42\% & 33\%\tabularnewline
 Worse  & 0\% & 0\% & 0\% & 0\% & 0\% & 17\%\tabularnewline
\hline 
\end{tabular}
\end{table}

\section{Comparative Study with Standard Global Optimization Solvers\label{sec:Comparative-Study}}

The goal of BO is to reduce the computational costs required to optimize
an expensive-to-evaluate function $f$, by reducing the number of
function evaluations. This section corroborates the claim by presenting
a proof-of-concept study recording the number of function evaluations
required to reach a log\textsubscript{10} distance to the true global
optimum equal to $-6$. In this experiment we used the objective function
CSF, shown in the top left panel of Figure \ref{fig:test_functions},
and compared ScaledEI with a range of algorithms widely used, for example, by
applied mathematicians and engineers:
\begin{enumerate}
\item Genetic Algorithm \citep{Goldberg1989,Conn1991,Conn1997};
\item Global Search \citep{Ugray2007};
\item Simulated Annealing \citep{Ingber1996};
\item Particle Swarm \citep{Mezura-Montes2011,Pedersen2010};
\item Multi Start (10 random starting points) \citep{Ugray2007,Glover1998};
\item Pattern Search \citep{Audet2002,Abramson2009}.
\end{enumerate}
We used the implementation found in MATLAB's Global
Optimization Toolbox\footnote{\url{https://uk.mathworks.com/products/global-optimization.html}}, with the default automatic settings for each algorithm.
Each optimization was repeated 15 times, using different random number generator seeds. 

\begin{figure}[tb]
\begin{centering}
\includegraphics[scale=0.5, trim={1cm 0 2cm 1.5cm}, clip]{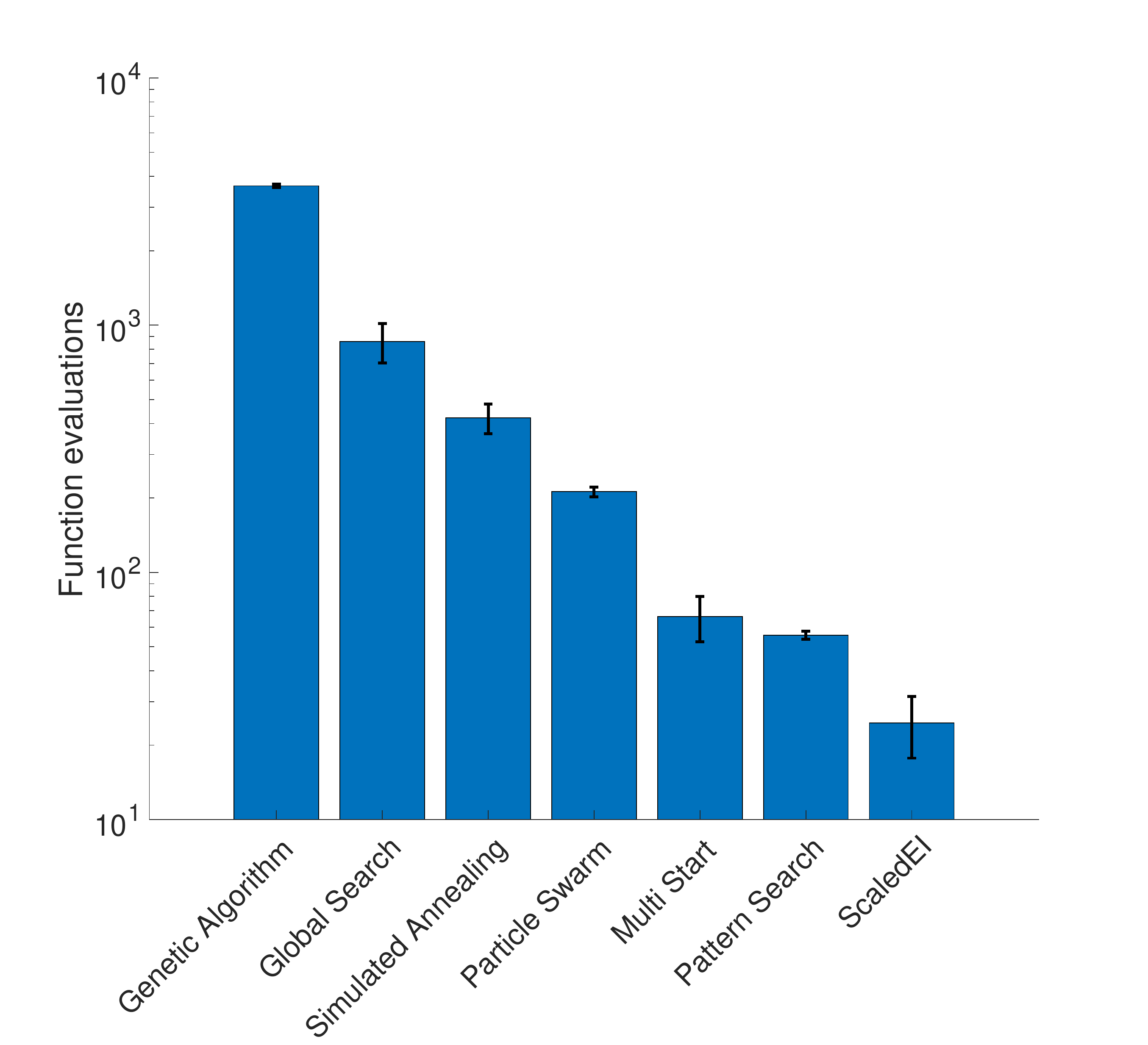}
\par\end{centering}
\caption{\textbf{Number of function evaluations required to reach a log}\protect\textsubscript{\textbf{10}}\textbf{
distance to the true global optimum (in function space) equal to $\boldsymbol{-6}$.}\label{fig:function_evaluations_csf}}
\end{figure}

Figure \ref{fig:function_evaluations_csf}
shows, for each optimization algorithm, the average number of function
evaluations (over the 15 random number seeds) required to reach a
log\textsubscript{10} distance of $-6$, while the error bars show
plus or minus one standard error of the mean. 
The Genetic Algorithm,
requiring between $10^{3}$ and $10^{4}$ function evaluations, is
the least suitable algorithm for expensive objective functions. Then
follow Global Search, which requires in the order of $10^{3}$ evaluations,
Simulated Annealing and Particle Swarm, both requiring between $10^{2}$
and $10^{3}$ evaluations. The Multi Start and Pattern Search solvers
rank as the most efficient ones, in terms of the number of function evaluations, but they are clearly outperformed
by ScaledEI. 
This comes as no surprise as, by construction, BO algorithms use all the information available from previous function evaluations to internally maintain a surrogate model of the objective function and infer its geometric properties in order to recommend the next candidate point.

\section{Application to In Silico Medicine\label{sec:Pulmonary}}

Chronic pulmonary arterial hypertension (PH), i.e. high blood pressure in the pulmonary  circulation, is often referred to as a ``silent killer'' and is a disease of the
small pulmonary arteries. It can lead to irreversible changes in the
pulmonary vascular structure and function, increased pulmonary vascular
resistance, and right ventricle hypertrophy leading to right heart
failure \citep{Allen2014,Rosenkranz2015}.

For diagnosis and ongoing treatment and assessment, clinicians measure
blood flow and pressure within the pulmonary arteries. 
As opposed to blood pressure in the systemic circulation, measured using a sphygmomanometer,
blood pressure in the pulmonary circulation can only be measured using invasive techniques such as right heart catheterization.
Invasive techniques can lead to complications (internal bleeding, severe pain, thrombosis, etc.), for that reason, it is desirable to predict the blood pressure indirectly based on quantities that can be measured non-invasively. 
Furthermore, data about healthy patients are not available due to ethical reasons. This
section uses a partial differential equations (PDEs) model of the
pressure and flow wave propagation in the pulmonary circulation under
normal physiological and pathological conditions, introduced by \citet{Qureshi2014}
and also studied by \citet{Noe2017}. The goal is to use the novel
ScaledEI Bayesian optimization algorithm introduced in (\ref{eq:scaled_ei})
and the pulmonary circulation model cited above to infer indicators
of pulmonary hypertension risk which could be used by clinicians to
inform their diagnosis instead of taking invasive measurements.

The PDEs depend on various physiological parameters, related e.g.
to blood vessel geometry, vessel stiffness and fluid dynamics. These
parameters, which would give important insights into the status of a patient's pulmonary circulatory system,  can typically not
be measured in vivo and hence need to be inferred indirectly from
the observed blood flow and pressure distributions. In principle,
this is straightforward. Under the assumption of a suitable noise
model, the solutions of the PDEs define the likelihood of the data,
and the parameters can then be inferred in a maximum likelihood sense.
However, a closed-form solution of the maximum likelihood equations
is not available, which calls for an iterative optimization procedure.
Since a closed-form solution of the PDEs is not available either,
each optimization step requires a numerical solution of the PDEs.
This is computationally expensive, especially given that the likelihood
function is typically multi-modal, and the optimization problem is
NP-hard. Minimization of the residual sum of squares (negative log
likelihood) hence calls for Bayesian optimization in order to reduce
the computational costs of the inference. The estimated parameters
of the PDEs will give clinicians insights into the patient-specific
vessel structure that would not be obtainable in vivo such as vessel
stiffnesses, a primary indicator of hypertension.

\subsection{The Pulmonary Circulation Model}

\begin{figure}[tb]
\centering{}\includegraphics[scale=0.5]{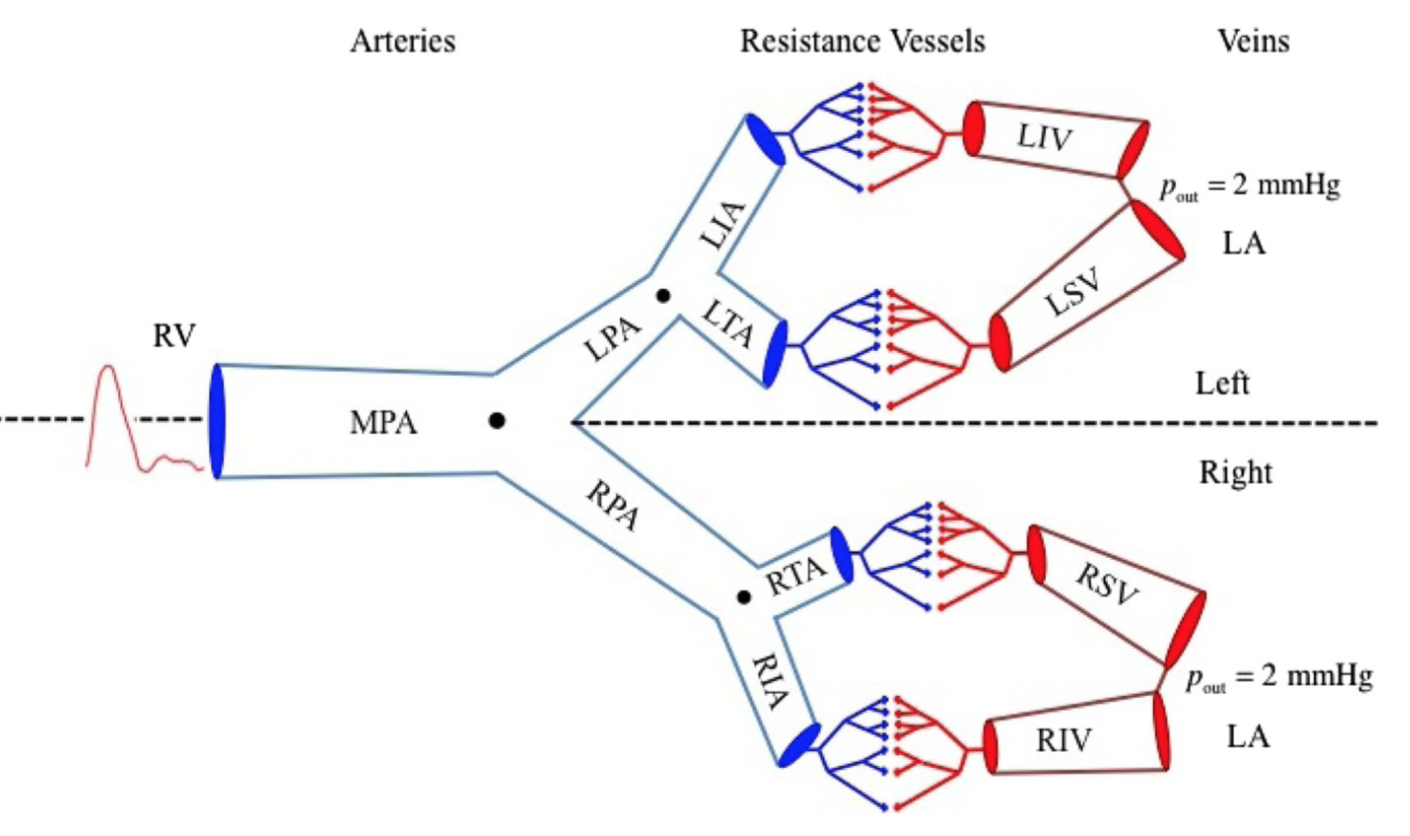} \caption{\textbf{Schematic of the pulmonary circulation consisting of large
arteries, arterioles, venules and large veins from~\citet{Qureshi2014}.}
Seven large arteries are considered in this model, i.e. the main pulmonary
artery (MPA), the left (LPA) and right (RPA) pulmonary arteries, the
left interlobular artery (LIA), the left trunk artery (LTA), the right
interlobular artery (RIA), and the right trunk artery (RTA). The four
terminal arteries LIA, LTA, RIA, and RTA are connected to four large
veins, i.e. the left inferior vein (LIV), left superior vein (LSV),
right inferior vein (RIV), and right superior vein (RSV), via structured
trees of resistance vessels. \label{fig:pc_vessels}}
\end{figure}

In the model of the pulmonary circulation by \citet{Qureshi2014},
seven large arteries and four large veins are modelled explicitly,
while the smaller vessels are represented by structured trees (Figure~\ref{fig:pc_vessels}).
A magnetic resonance imaging (MRI) based measurement of the right
ventricular output provides the inlet flow for the system.

The large arteries and veins are modelled as tapered elastic tubes,
and the geometries are based on measurements of proximal and distal
radii and vessel lengths. The cross-sectional area averaged blood
flow and pressure are predicted from a non-linear model based on the
incompressible Navier\textendash Stokes equations for a Newtonian
fluid. The small arteries and veins are modelled as structured trees
at each end of the terminal large arteries and veins to mimic the
dynamics in the vascular beds. With a given parent vessel radius $r_{p}$,
the daughter vessels are scaled linearly with radii $r_{d_{1}}=\alpha r_{p}$
and $r_{d_{2}}=\beta r_{p}$, where $\alpha$ and $\beta$ are the
scaling factors. The vessels bifurcate until the radius of each terminal
vessel is smaller than a given minimum $r_{\min}$. The radius relation
at bifurcations is:
\begin{equation}
r_{p}^{\xi}=r_{d_{1}}^{\xi}+r_{d_{2}}^{\xi},\quad2.33\leq\xi\leq3.0,\label{eq:xi}
\end{equation}
where the exponent $\xi=2.33$ corresponds to laminar flow, $\xi=3.0$
corresponds to turbulent flow~\citep{Olufsen1999}, $p$ represents
the parent vessel, and $d_{1}$ and $d_{2}$ represent the daughter
vessels. Given the area ratio $\eta=(r_{d_{1}}^{2}+r_{d_{2}}^{2})/{r_{p}^{2}}$
and the asymmetry ratio $\gamma=(r_{d_{2}}/r_{d_{1}})^{2}$, the scaling
factors $\alpha$ and $\beta$ satisfy $\alpha=(1+\gamma^{\xi/2})^{-1/\xi}$
and $\beta=\alpha\sqrt{\gamma}$. The parameters, $\xi$, $\gamma$,
$r_{\min}$ and a given root radius $r_{0}$, determine the size and
density of the structured tree. The cross-sectional area averaged
blood flow and pressure in these small arteries and veins are computed
from the linearized incompressible axisymmetric Navier\textendash Stokes
equations~\citep{Qureshi2014}.

For each large vessel, the pressure and flow are modelled as the solution
of the one dimensional Navier\textendash Stokes equation \citep{Olufsen2012}.
It comprises two equations which ensure \emph{conservation of volume
and momentum,} and a third equation of state, relating \emph{pressure
and cross-sectional area}. Let $x$ denote the distance along a given
vessel, $t$ represent time, $p(x,t)$ the pressure, $q(x,t)$ the
volumetric flow along any given vessel, $A(x,t)$ the corresponding
cross-sectional area, $\rho$ a constant representing the density
of the blood, $\nu$ a constant representing kinematic viscosity,
$\delta$ (constant) the boundary layer thickness, and $r(x,t)$ the
radius of the given vessel. Conservation of volume and momentum is
satisfied by:

\begin{equation}
\frac{\partial q}{\partial x}+\frac{\partial A}{\partial t}=0,\qquad\frac{\partial q}{\partial t}+\frac{\partial}{\partial x}\left(\frac{q^{2}}{A}\right)+\frac{A}{\rho}\frac{\partial p}{\partial x}=-\frac{2\pi\nu r}{\delta}\frac{q}{A}.\label{eq:conservation-volume-momentum}
\end{equation}
The constitutive law linking pressure and cross sectional area is
given by:

\begin{equation}
p(x,t)-p_{0}=\frac{4}{3}\frac{Eh}{r_{0}}\left(1-\sqrt{\frac{A_{0}}{A}}\right),\label{eq:pressure-area}
\end{equation}
where $p_{0}$ denotes the external pressure, $E$ is Young's modulus,
$h$ the vessel wall thickness and $r_{0}$ the vessel radius when
$p(x,t)=p_{0}$. The unstressed vessel area is obtained as $A_{0}=\pi r_{0}^{2}$.
The term $Eh/r_{0}$ in (\ref{eq:pressure-area}) describes the elastic
properties of a vessel's wall, and hence represents a parameter that
controls the system compliance. This will be simply denoted by $f_{\mathrm{L}}$
in the large vessels and by $f_{\mathrm{S}}$ in the small vessels.

\begin{figure}[tb]
\begin{centering}
\includegraphics[width=0.5\textwidth]{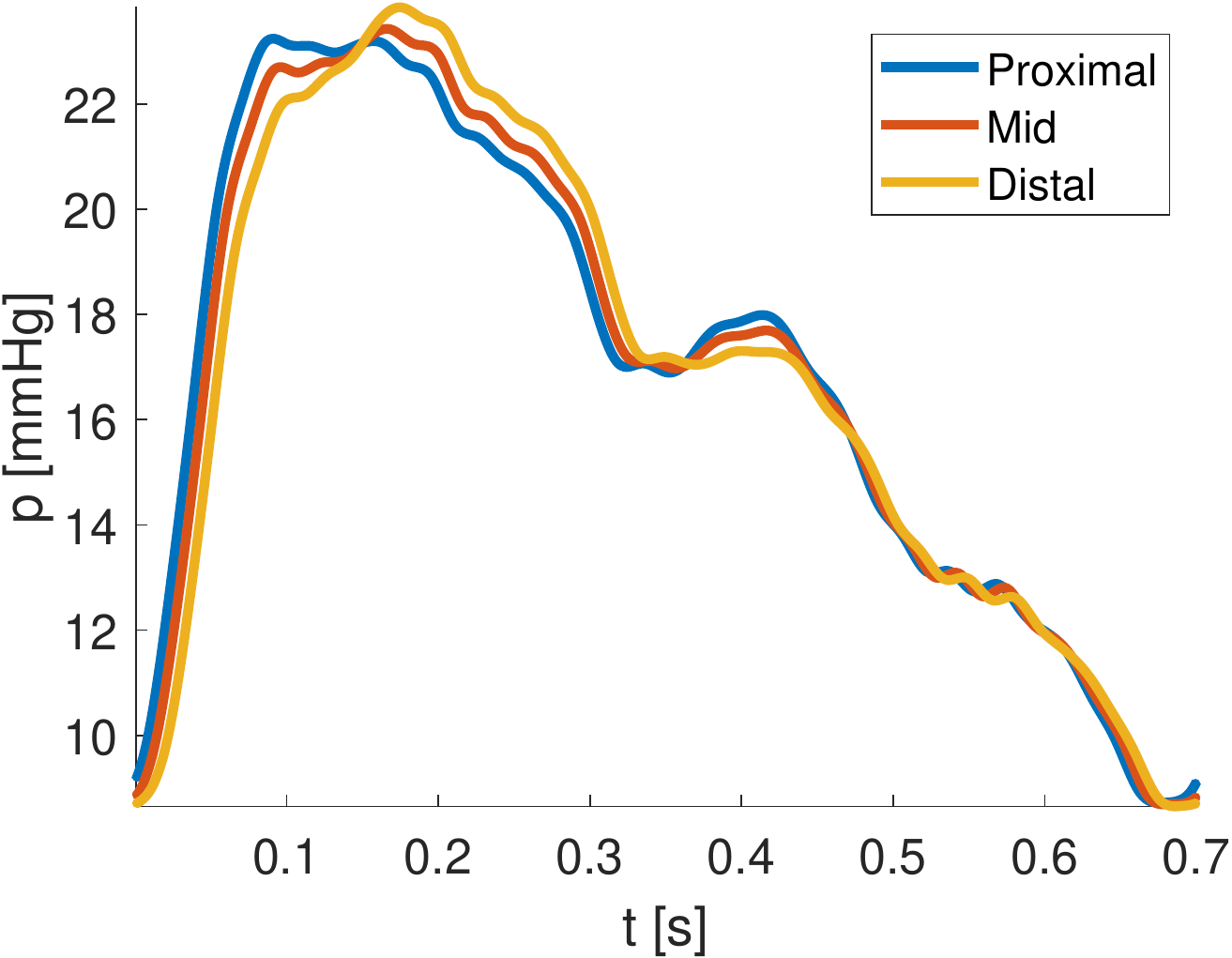}\includegraphics[width=0.5\textwidth]{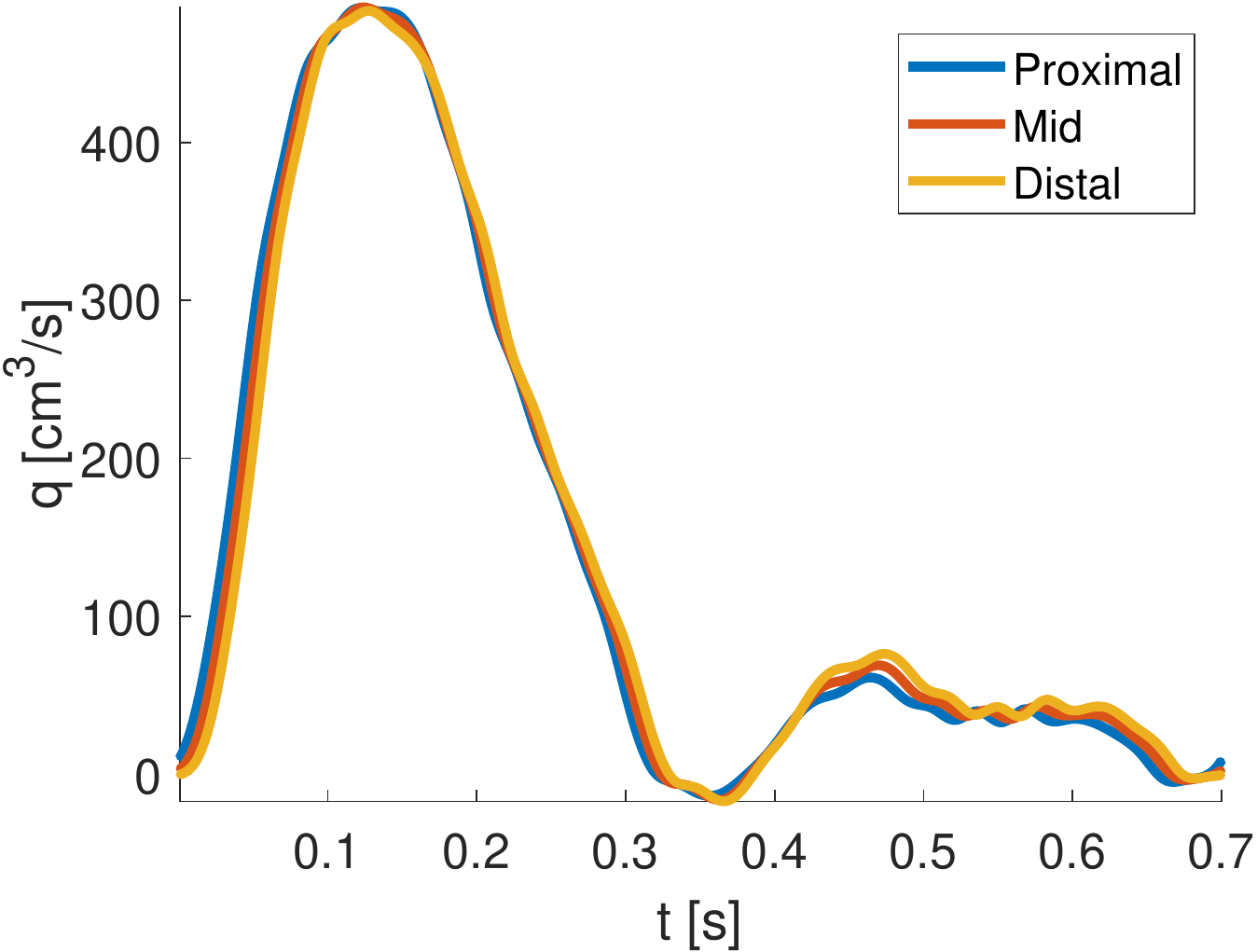}
\par\end{centering}
\caption{\textbf{Simulated pressure (left) and flow (right) over time, at three different
locations along the MPA.}\label{fig:simulated-p-q}}
\end{figure}

In the small vessels, similarly to the large ones, three equations
determine the flow, pressure and area of each vessel in the structured
tree. \citealt{Olufsen2012} however, notice that in small vessels
the nonlinear effects are small, effectively allowing for linearization
of the constitutive equations. The full system of PDEs is presented
in \citet{Qureshi2014}, and its numerical solution, which depends
on various physiological parameters, will henceforth be referred to
as \emph{simulation}. 
Figure \ref{fig:simulated-p-q} shows the simulated
pressure (left) and flow (right) curves over time at three different
locations in the main pulmonary artery (MPA). 

Particular interest lies in the estimation of
the parameter $\xi$, because low values are indicative of the clinically
relevant problem of \emph{vascular rarefaction}, which is a well-known
finding in patients suffering from pulmonary hypertension, and represents
the condition of having fewer blood vessels per tissue volume \citep{Feihl2006}.
Estimation of $\xi$ is performed in the range $2.33\leq\xi\leq3$.
Other relevant parameters of interest for clinical diagnosis are the
stiffness parameters in the large and small vessels, $f_{\mathrm{L}}$
and $f_{\mathrm{S}}$ respectively, with bounds $f_{\mathrm{L}}\in[1.33\times10^{5},5.33\times10^{5}]$
and $f_{\mathrm{S}}\in[2.66\times10^{4},1.066\times10^{5}]$ as in \cite{Noe2017}. Increased
vessel stiffness is a major cause of pulmonary hypertension. During
systole, a compliant artery expands to accommodate for the inflow,
while it recoils during diastole to promote forward flow. As the capacity
of an artery is limited, the pressure increases during systole and
is partially maintained during diastole by the rebounding of the expanded
arterial walls. When the stiffness is increased, the cushioning function
of the vessel is compromised, leading to a higher systolic and a lower
diastolic pressure. All remaining model parameters are fixed to biologically
relevant value from the literature as in \citet{Qureshi2014}.

\subsection{Bayesian Optimization with Hidden Constraints}
Mathematical models often rely on simplifying assumptions about the underlying system. When these assumptions do not hold, the model can return a failure instead of a numerical simulation.
In the considered model of the human pulmonary circulation, for some specific settings of the PDE parameters this is indeed the case. However, the regions in the parameters space that lead to failures are unknown a priori.
A problem is said to contain \emph{hidden constraints} if a requested function value may turn out not to be obtainable. This is different from the case where an output value is obtained but deemed not valid.

In this case the optimization of the function must be performed hand-in-hand with a sequential learning of the domain areas leading to numerical failures. This requires a modification of Algorithm \ref{alg:bo}, as in \cite{Gelbart2014}. The idea is that, along with each function evaluation $y_i = f(\xb_i)$, we also keep track of the failure or success of the query in an auxiliary variable $h_i = h(\xb_i)$. The convention used in our work for the binary variable $h(\xb)$ is to take the value $1$ in case of a failure and $-1$ for a successful evaluation. Hence, we name $h \in \{-1, 1\}$ the failure indicator.
The initial design will consist of triples $(\xb_i, y_i, h_i)$ for $i = 1, \dots, \ninit$. The next step consists in obtaining two GP models: 
\begin{enumerate}[label=(\alph*)]
\item a GP model of the objective function, using the $(\xb_i, y_i)$ pairs;
\item a GP model of the failures, using the $(\xb_i, h_i)$ pairs. 
\end{enumerate}
Model (a) represents a standard GP regression as described in Section \ref{subsec:gaussian_processes}, while model (b) requires predicting the posterior class probabilities of $h$ for a new input $\xb$, given a set of training data.
For the GP classification with logit or probit link function, the class posterior probability is analytically intractable. Different approximations have been proposed in order to predict binary-valued outcomes, for example iterative procedures like Expectation Propagation (EP), Laplace approximation (LA) or the simpler label regression (LR) approach. 

Following the experiments and recommendations of \cite{Kuss2006}, which suggest that label regression works surprisingly well in practice and with a lower error rate than the competing methods in high dimensions, we apply LR in order to build a model of the simulation failures, ignoring the binary nature of the variable $h$ in favor of a quicker runtime and a closed-form posterior.
Any approximation involving extra iterative procedures could cause the modelling and point selection time to be higher than simply evaluating the objective function. In other words, we would spend more time in modelling the function rather than evaluating it. While this can be arguably acceptable for really expensive simulators, it would not be computationally optimal for the presented application. A discussion of the computational costs of the pulmonary circulation model under study can be found in Section \ref{subsec:pc_estimation}.

Denote the failure indicator model as $h(\xb) \sim \GP{\hat{h}(\xb)}{s_h(\xb, \xb')}$, obtained by applying the formulas summarized in Section \ref{subsec:gaussian_processes} to the $\{ (\xb_i, h_i) \}$ data. By the marginalization property of GPs, at point $\xb$ the random variable $h(\xb)$ follows a $\N(\hat{h}(\xb), s_h^2(\xb))$ distribution.
As failures are labelled as $1$ and successful evaluations as $-1$, by taking the probability of the Gaussian random variable being less than $0$ we obtain an indication of the probability of a successful evaluation (no failure):
\begin{equation}
\mathbb{P}\{ h(\xb) = -1 \} = \Phi(0 \mid \hat{h}(\xb), s^2_h(\xb)).
\label{eq:prob_success}
\end{equation}
This probability can then be used to weight the score that any acquisition function assigns to a point in the domain as follows \citep{Gelbart2014}:
\begin{align}\label{eq:hcw_acquisition}
a_n^*(\xb) &= a_n(\xb) \times \mathbb{P}\{ h(\xb) = -1 \} \\
&= a_n(\xb) \times \Phi(0 \mid \hat{h}(\xb), s^2_h(\xb)). \nonumber
\end{align}
We refer to $a_n^*(\xb)$ as the hidden-constraints-weighted acquisition function. The algorithm then proceeds normally by choosing the next query point as the point maximizing $a_n^*(\xb)$. A pseudocode summary can be found in Algorithm \ref{alg:bo_errors}. From \eqref{eq:hcw_acquisition} we see that both models (a) and (b) are required in order to compute the hidden-constraints-weighted acquisition function at each iteration of the BO algorithm.

\begin{algorithm}[tb]
   \caption{Bayesian optimization with hidden constraints.\label{alg:bo_errors}}
\begin{algorithmic}[1]
   \STATE \textbf{Inputs:} \\
   Initial design and corresponding failure labels: $\D_{\ninit} = \{ (\xb_i, y_i, h_i) \}_{i=1}^{\ninit}$ \\
   Budget of $ \nmax $ function evaluations
   \item[]
   \FOR{$n = \ninit$ \textbf{to} $\nmax - 1$}
   \STATE Update the objective GP: $f(\xb) \mid \D_{n} \sim \GP{\fhat(\xb)}{s(\xb,\xb')}$
   \STATE Update the failure GP: $h(\xb) \mid \D_{n} \sim \GP{\hat{h}(\xb)}{s_h(\xb,\xb')}$
   \STATE Compute the acquisition function: $ a_n^*(\xb) = a_n(\xb) \times \Phi(0 \mid \hat{h}(\xb), s_h^2(\xb))$
   \STATE Solve the auxiliary optimization problem: $\xnext = \argmax_{\xb \in \bb{X}} a_n^*(\xb)$
   \STATE Query $f$ at $\xnext$ to obtain $\fnext$ and $h_{\next}$
   \STATE Augment data: $\D_{n + 1} = \D_{n} \cup \{ \xnext, \fnext, h_{\next} \}$
   \ENDFOR
   \item[]
   \STATE \textbf{Return:} \\
   Estimated minimum: $f_\mathrm{min} = \min(y_1, \dots, y_{\nmax})$ \\
   Estimated point of minimum: $\boldsymbol{x}_\mathrm{min} = \argmin(y_1, \dots, y_{\nmax})$
\end{algorithmic}
\end{algorithm}

\begin{figure}[tb]
\begin{centering}
\includegraphics[scale=0.5]{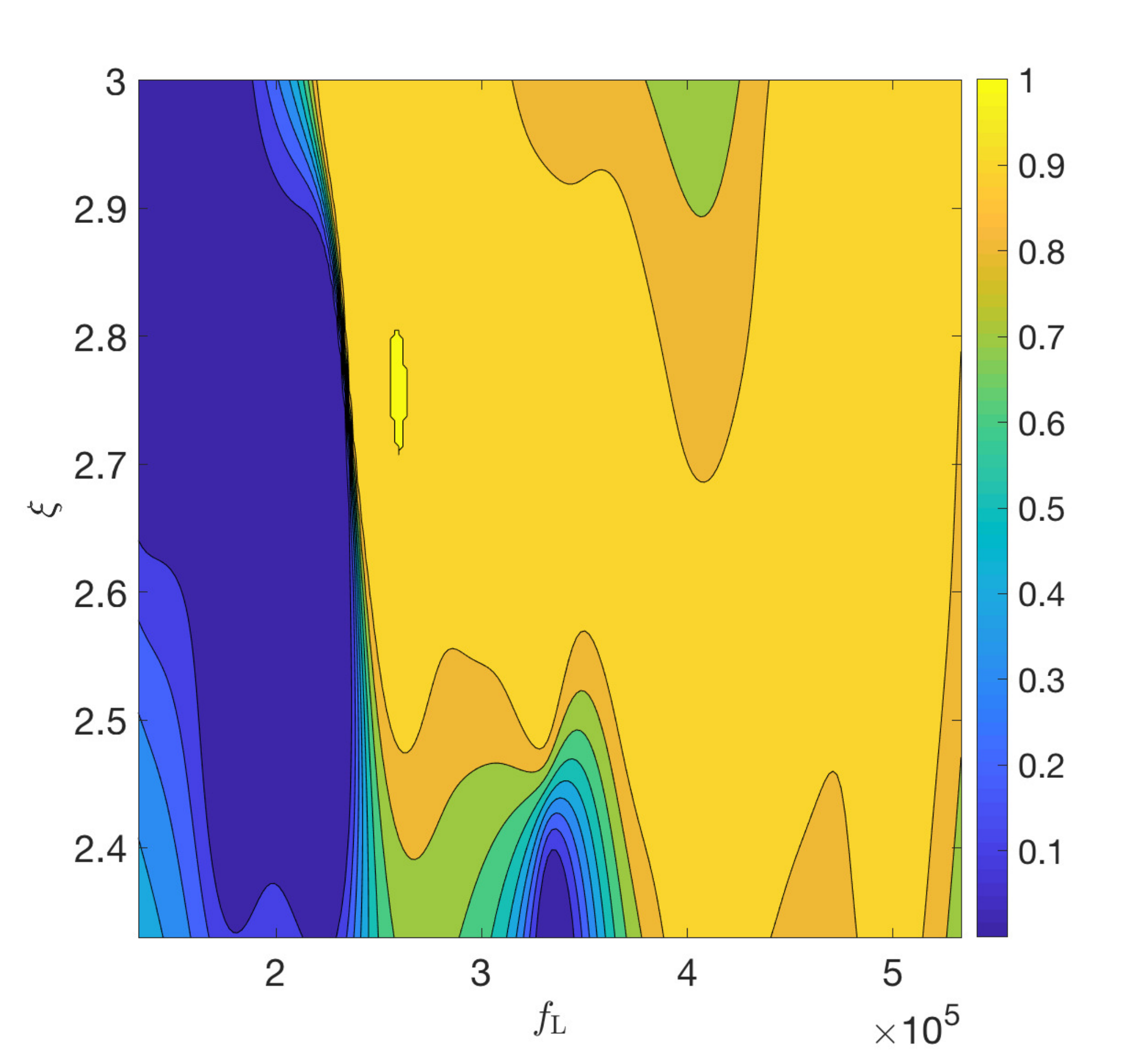}
\par\end{centering}
\caption{\textbf{Learned probability of a successful simulation in the PDEs model of the human pulmonary circulation.} A score near 1 indicates a high chance of a successful simulation, while a score near 0 indicates a high probability of failure.\label{fig:prob_no_failure_2d}}
\end{figure}

At termination, the learned failure GP model can be used to obtain insights into the regions in the parameter domain leading to failure. 
For plotting purposes, in Figure \ref{fig:prob_no_failure_2d} we show the probability of a successful evaluation (no failure), $\mathbb{P}\{ h(\xb) = -1 \} = \Phi(0 \mid \hat{h}(\xb), s^2_h(\xb))$, for the 2D parameter space $\xb = (f_\mathrm{L}, \xi)^\T$.
The figure shows how it is possible to have regions of failure inside a larger area of successful simulations.

\subsection{Estimation of the Model Parameters\label{subsec:pc_estimation}}

In the computational model of the pulmonary circulation, denoted by
$\mathsf{m}$, a forward simulation for fixed parameters takes around 23 seconds of CPU
time\footnote{On a Dell Precision R7610 workstation with dual 10core Intel Xeon CPU with hyper-threading and 32GB RAM.}. The data collected by clinicians typically include pressure
and flow measurements only from the midpoints of the 11 large vessels.
In light of this, we refer to a \emph{simulation} as the $22$-dimensional
vector $\boldsymbol{y}=\mathsf{m}(\boldsymbol{q})$ containing pressure
and flow measurements at the midpoint location of each of the 7 large
arteries and 4 large veins, for a given vector of PDE parameters $\boldsymbol{q}$.
Given the costs of a single function evaluation, parameter estimation
comes at substantial computational demands as standard global optimization
algorithms require a large number of forward simulations (for example
see Figure \ref{fig:function_evaluations_csf}). Motivated by real-time
decision making, we want to reduce the computational time required
to estimate the PDE parameters by keeping the number of function evaluations
as low as possible. To do so, we tackle this problem by using Bayesian
optimization with hidden constraints and the novel ScaledEI acquisition function introduced
in (\ref{eq:scaled_ei}).

Let $\boldsymbol{q}=(f_{\mathrm{L}},f_{\mathrm{S}},\xi)^\T$ denote the
three parameters of relevance for the diagnosis of pulmonary hypertension.
 The simulated pressure and flow data $\boldsymbol{y}_{0}$ are obtained
by a forward simulation of the computational model at the vector of
parameters $\boldsymbol{q}_{0}=(2.6\times10^{5},5\times10^{4},2.76)^\T$,
assumed to be the underlying truth, and the data are then corrupted by i.i.d. additive Gaussian noise with a signal-to-noise ratio (SNR) of 10db. 
Pretending that the true parameter vector $\boldsymbol{q}_{0}$
is unknown, interest lies in its estimation from the noisy observations
$\boldsymbol{y}_{0}$. This is to present a proof-of-concept study carried out on simulated data,  for which it is possible to assess the inference as the gold-standard is known, but with the objective to ultimately apply it to real data and move it into the clinic. 
For i.i.d. additive Gaussian noise, the residual sum of squares is proportional to the negative log likelihood. Hence, the objective function considered in this study
is the squared L2 loss, or residual sum of squares, between a simulation
$\mathsf{m}(\boldsymbol{q})$ and the data $\boldsymbol{y}_{0}$:
\begin{equation}
\ell(\boldsymbol{q})=\|\mathsf{m}(\boldsymbol{q})-\boldsymbol{y}_{0}\|^{2}.\label{eq:pc-objective-fcn}
\end{equation}
Since each evaluation of the squared loss $\ell$ involves an expensive
forward simulation from the PDE model $\mathsf{m}(\boldsymbol{q})$,
also each evaluation of $\ell$ will be expensive. We estimate the
parameters $\boldsymbol{q}=(f_{\mathrm{L}},f_{\mathrm{S}},\xi)^\T$ by
minimizing the squared loss using Bayesian optimization with hidden constraints and the ScaledEI
acquisition function, with the following notational correspondence
in Algorithm~\ref{alg:bo_errors}:
\begin{itemize}
\item Objective function: $f(\boldsymbol{x})\equiv\ell(\boldsymbol{q})$;
\item Input: $\boldsymbol{x}_{i}\equiv\boldsymbol{q}_{i}$;
\item Output: $y_{i}\equiv\ell_{i}$.
\end{itemize}
In this experiment we set a priori the maximum budget of function
evaluations to $\nmax=500$. We repeated the optimization of $\ell$
using fifteen different random number seeds.
Let $\ell_{\min}=\min(\ell_{1},\dots,\ell_{n})$ denote the minimum
observed objective (current best function value) at iteration $n$.
\begin{figure}[tb]
\begin{centering}
\includegraphics[width=\textwidth]{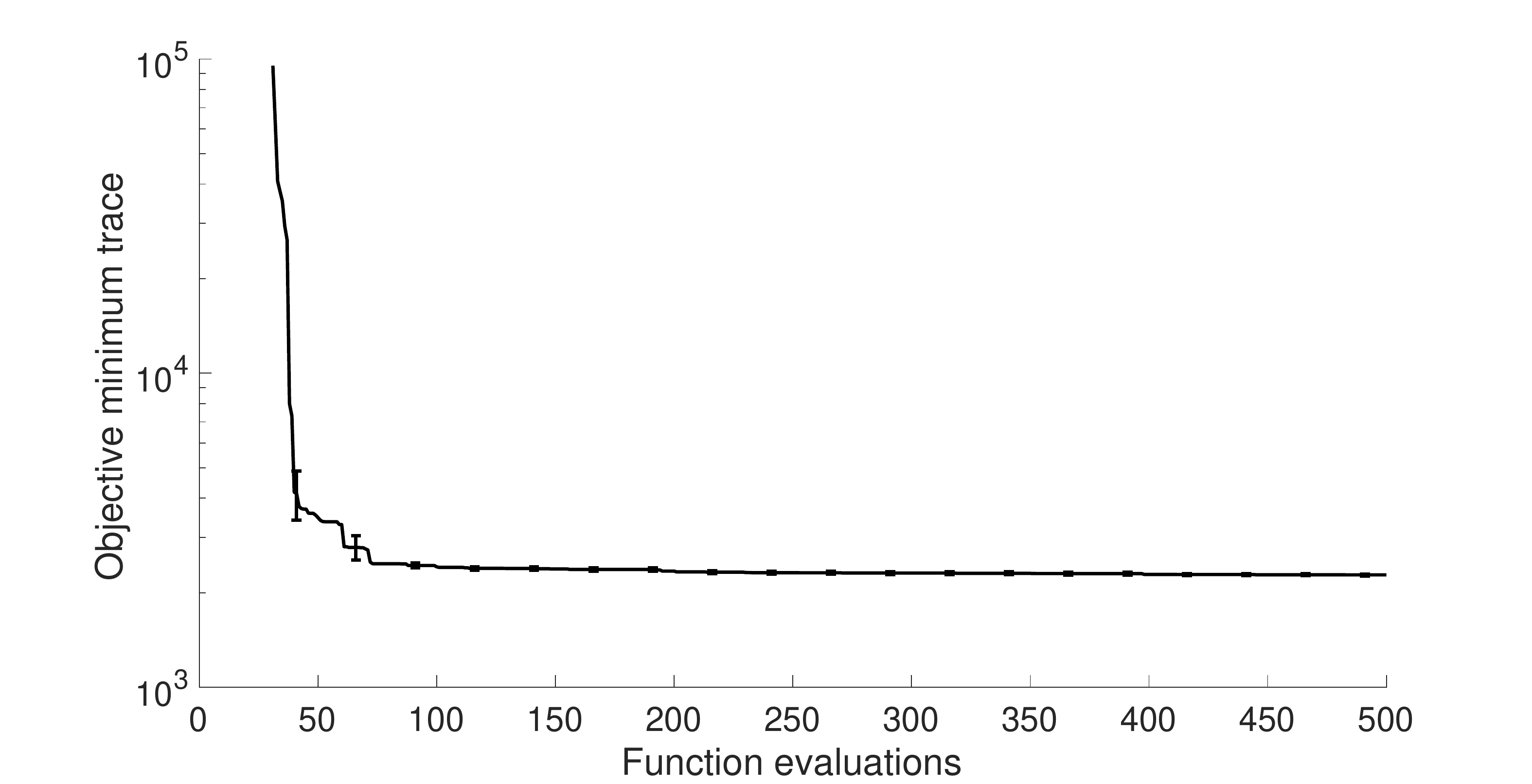}
\par\end{centering}
\caption{\textbf{Optimization of the squared L2 loss for the pulmonary circulation model using the ScaledEI BO algorithm.}
The plot shows the minimum observed squared L2 loss trace, averaged over the 15 Latin hypercube design instantiations, plus or minus one standard error, as function of the number of function evaluations. \label{fig:pc-obj-min-trace}}
\end{figure}
Figure \ref{fig:pc-obj-min-trace} shows the average objective minimum
trace, over the fifteen designs, as function of the number of function
evaluations $n$ $(n=1,\dots,\nmax)$. The error bars represent plus
or minus one standard error of the mean. When the maximum budget of
function evaluations is exceeded, the BO algorithm stops by returning
the estimated objective minimum, $\ell_{\min}$, and the point $\boldsymbol{q}_{\min}$
at which the minimum is attained. The vector $\boldsymbol{q}_{\min}$
represents the estimate of the true but unknown $\boldsymbol{q}_{0}$.

\begin{table}[tb]
\caption{\textbf{The PDE parameters underlying the simulated data (Truth) and
the estimated parameters (Estimate). }Mean and standard error over
the 15 design instantiations. \label{tab:PDE-parameter-estimates}}

\centering{}%
\begin{tabular}{cccccc}
\hline 
 & \multirow{2}{*}{Truth} & \multicolumn{2}{c}{Estimate $(n=500)$} & \multicolumn{2}{c}{Estimate $(n=100)$}\tabularnewline
 &  & Mean & Std. Err. & Mean & Std. Err.\tabularnewline
\hline 
$f_{\mathrm{L}}$ & $2.6\times10^{5}$ & $2.6005\times10^{5}$ & $189.3129$ & $2.599\times10^{5}$ & $286.5552$\tabularnewline
$f_{\mathrm{S}}$ & $50000$ & $50003$ & $35.0108$ & $50038$ & $55.9675$\tabularnewline
$\xi$ & $2.76$ & $2.7603$ & $0.0002$ & $2.76$ & $0.0004$\tabularnewline
\hline 
\end{tabular}
\end{table}

Table \ref{tab:PDE-parameter-estimates} shows the average and standard
error, over the 15 repetitions, of the estimated point of minimum
$\boldsymbol{q}_{\min}$ at iteration $n=\nmax$, next to the truth
$\boldsymbol{q}_{0}$ used to generated the data. Considering that the data
were corrupted by noise, the estimation (Mean) is accurate, and with reasonably small uncertainty (Std. Err.) given the scale of each parameter. Furthermore, each element of the true parameter vector $\boldsymbol{q}_{0}$ lies inside the 95\% confidence interval obtained as the Mean plus or minus two Std. Err. in Table~\ref{tab:PDE-parameter-estimates}. We
notice that the curve in Figure~\ref{fig:pc-obj-min-trace} is approximately
flat after 100 function evaluations. For this reason, we could have
effectively stopped at approximately between 100 and 200 iterations,
without a substantial loss in accuracy, while reducing the overall
computational time from 3 hours to less than 1 hour. In Table \ref{tab:PDE-parameter-estimates} we also report the estimated parameters stopping at $n=100$ function evaluations only, where one run of the optimization takes approximately 30 minutes, compared to the 3 hours required for $n=500$. These timings can be used for in-clinic decision support systems of practical relevance. However, for a standard optimization algorithm requiring, for example, $10^4$ function evaluations, one run of the algorithm would have taken approximately 3 days, making traditional algorithms not suitable for in-clinic applications.

\section{Conclusions\label{sec:conclusions}}

We have proposed a new acquisition function for Bayesian optimization
(BO) which falls into the class of improvement-based policies (class 2), summarized in Section~\ref{subsec:bayesian_optimization}.
It is based on a random variable, called \emph{Improvement, }defined
in~(\ref{eq:improvement}), which quantifies the improvement on the
incumbent optimum. We have discussed that the established \emph{Expected
Improvement} acquisition function~(\ref{eq:expected_improvement})
does not account for the uncertainty in the \emph{Improvement} random
variable, which conveys information about our confidence in its value.
We have overcome this problem by deriving the variance of \emph{Improvement}
in~(\ref{eq:variance_of_improvement}) and using it to derive a new
acquisition function, referred to as ScaledEI, which is the ratio
of the \emph{Expected Improvement} and the standard deviation of \emph{Improvement},
see~(\ref{eq:scaled_ei}). The proposed acquisition function accounts
for another source of uncertainty, and the scaling factor plays a
role in both exploitational and explorational moves. By selecting
the point that maximizes the ScaledEI policy we effectively select
a point for which we expect, on average, a high degree of improvement
at high confidence.

We have evaluated the performance of the proposed acquisition function
on an extensive set of benchmark problems from the global optimization
literature. The test suite includes problems of different dimensionality,
varying from 1D to 10D, having multiple local minima and, additionally,
symmetries corresponding to multiple equivalent global minima.

The performance was evaluated in terms of the log\textsubscript{10}
distance (in function space) to the global optimum. The results indicate
that ScaledEI tends to perform as well as or better than the representative
set of state-of-the-art methods from the BO literature included in
our study. This suggests that by adopting a new search strategy that
explicitly combines the expected improvement with its estimated uncertainty,
we obtain a better trade-off between \emph{exploration} and \emph{exploitation}.
The result is a new competitive search strategy that does not only
compare favourably with other improvement-based alternatives (class
2), but also with optimistic (class 1) and information-theoretic (class
3) strategies.

We have then shown a proof-of-concept study that confirms the reduction
in the number of function evaluations required to optimize the CSF
function. The proposed ScaledEI algorithm was compared to a set of
widely used global optimization solvers, by reporting the number of
function evaluations required to reach a given tolerance on the $f$
value. The plot in Figure \ref{fig:function_evaluations_csf} confirms
that Bayesian optimization with the proposed ScaledEI acquisition function has indeed the lowest number of
function evaluations, since it uses a surrogate model of the objective
function to inform the next evaluation.

Our final contribution was a proof-of-concept study based on a PDE fluid dynamics model of the human pulmonary circulation, which is potentially relevant to precision medicine and non-invasive real-time diagnosis. The aim was to use the PDE model in order to give clinicians
three clear indicators of pulmonary hypertension, without going through
the invasive procedure of right heart catheterization. The three indicators
(large vessels stiffness, small vessels stiffness and density of the
structured tree, representing vascular rarefaction) are derived from
the parameters of the constitutive equations of the soft tissues,
which give pathophysiological insights that are very difficult to obtain in vivo. 
We have shown how to estimate the three parameters using the proposed ScaledEI method, introduced in Section \ref{sec:scaled_ei}. In particular, the estimates were obtained in a time frame that is suitable for in-clinic diagnosis and prognosis. As seen from Figure \ref{fig:function_evaluations_csf}, this goal would be more challenging to achieve with conventional non-Bayesian optimization routines. Hence, our combination of  the new ScaledEI method with the fledgling fluid dynamics model of the human pulmonary blood circulation system is an important first stepping stone on the pathway to an autonomous in silico clinical decision support system.

\newpage

\appendix\normalsize

\section{Appendix\label{sec:appendix}}

Let $f(\xb)\sim\GP{m(\xb)}{k(\xb,\xb')}$ be the GP prior on the objective
function. We recall that given data $\D_{n}=\{(\xb_{1},y_{1}),\dots,(\xb_{n},y_{n})\}$
the predictive distribution at $\xb$ is the marginal GP $f(\xb)\mid\D_{n}\sim\N(\fhat(\xb),s^{2}(\xb))$.
By standardization, $z(\xb)=\{f(\xb)-\fhat(\xb)\}/s(\xb)\sim\N(0,1)$.
Define the \emph{Improvement} at $\xb$ as the random variable $I(\xb)=\max\{\fmin-f(\xb),0\}$,
where $\fmin=\min(y_{1},\dots,y_{n})$ denotes the current best function
value at iteration $n$.

\subsection{Derivatives of Gaussian pdfs\label{subsec:derivatives_phi}}

Let $\phi(z)=(\sqrt{2\pi})^{-1}\exp(-z^{2}/2)$ be the standard Gaussian
pdf. Then, 
\begin{equation}
\phi'(z)=\frac{\d}{\d z}\phi(z)=\phi(z)\times\left(-\frac{1}{2}\times2z\right)=-z\phi(z).\label{eq:phi_d}
\end{equation}
The second derivative of the standard Gaussian pdf is: 
\begin{align}
\phi''(z) & =\frac{\d}{\d z}\phi'(z)\nonumber \\
 & =\frac{\d}{\d z}\left\{ -z\phi(z)\right\} \label{eq:phi_dd}\\
 & =-\phi(z)+(-z)(-z\phi(z))\nonumber \\
 & =(z^{2}-1)\phi(z).\nonumber 
\end{align}

\subsection{Derivation of the Probability of Improvement\label{subsec:derivation_pi}}

The \emph{Probability of Improvement} (PI) is the probability of the
event $\{I(\xb)>0\}$ or, equivalently, $\{f(\xb)<\fmin\}$:

\begin{align*}
\mathrm{PI}(\xb) & =\bb P\{I(\xb)>0\}\\
 & =\E1_{\{f(\xb)<\fmin\}}\\
 & =\bb P\{f(\xb)<\fmin\}\\
 & =\bb P\left\{ \frac{f(\xb)-\fhat(\xb)}{s(\xb)}<\frac{\fmin-\fhat(\xb)}{s(\xb)}\right\} \\
 & =\int_{-\infty}^{\frac{\fmin-\fhat(\xb)}{s(\xb)}}\phi(z)\d z\\
 & =\Phi\left(\frac{\fmin-\fhat(\xb)}{s(\xb)}\right),
\end{align*}
where $\phi(x\mid\mu,\sigma^{2})$ and $\Phi(x\mid\mu,\sigma^{2})$
represent the pdf and cdf of a $\N(\mu,\sigma^{2})$ distribution
evaluated at $x$ respectively. When $\mu=0$ and $\sigma^{2}=1$
we will simply write $\phi(x)$ and $\Phi(x)$ for brevity.

\subsection{Derivation of the Expected Improvement\label{subsec:derivation_ei}}

Define $u=\{\fmin-\fhat(\xb)\}/s(\xb)$. The expected value of the
random variable \emph{$I(\xb)$} is the \emph{Expected Improvement}
(EI) acquisition function: 

\begin{align*}
\mathrm{EI}(\xb) & =\E[I(\xb)]\\
 & =\E[\max\{\fmin-f(\xb),0\}]\\
 & =\E\left[\{\fmin-f(\xb)\}1_{\{f(\xb)<\fmin\}}\right]\\
 & =\int_{-\infty}^{\infty}\{\fmin-y\}1_{\{y<\fmin\}}\phi(y\mid\fhat(\xb),s^{2}(\xb))\d y\\
 & =\int_{-\infty}^{\fmin}\{\fmin-y\}\phi(y\mid\fhat(\xb),s^{2}(\xb))\d y\\
 & =\int_{-\infty}^{u}\{\fmin-(\fhat(\xb)+s(\xb)z)\}\phi(z)\d z\\
 & =\int_{-\infty}^{u}\{\fmin-\fhat(\xb)-s(\xb)z\}\phi(z)\d z\\
 & =\{\fmin-\fhat(\xb)\}\int_{-\infty}^{u}\phi(z)\d z-s(\xb)\int_{-\infty}^{u}z\phi(z)\d z\\
 & =\{\fmin-\fhat(\xb)\}\Phi\left(u\right)+s(\xb)\phi\left(u\right)\\
 & =s(\xb)\left\{ u\Phi(u)+\phi(u)\right\} .
\end{align*}

\subsection{Mat\'{e}rn 5/2 kernel\label{subsec:Matern-5/2-kernel}}

\begin{figure}[tbhp]
\begin{centering}
\includegraphics[trim={1cm 1cm 1cm 1cm},clip,width=\textwidth]{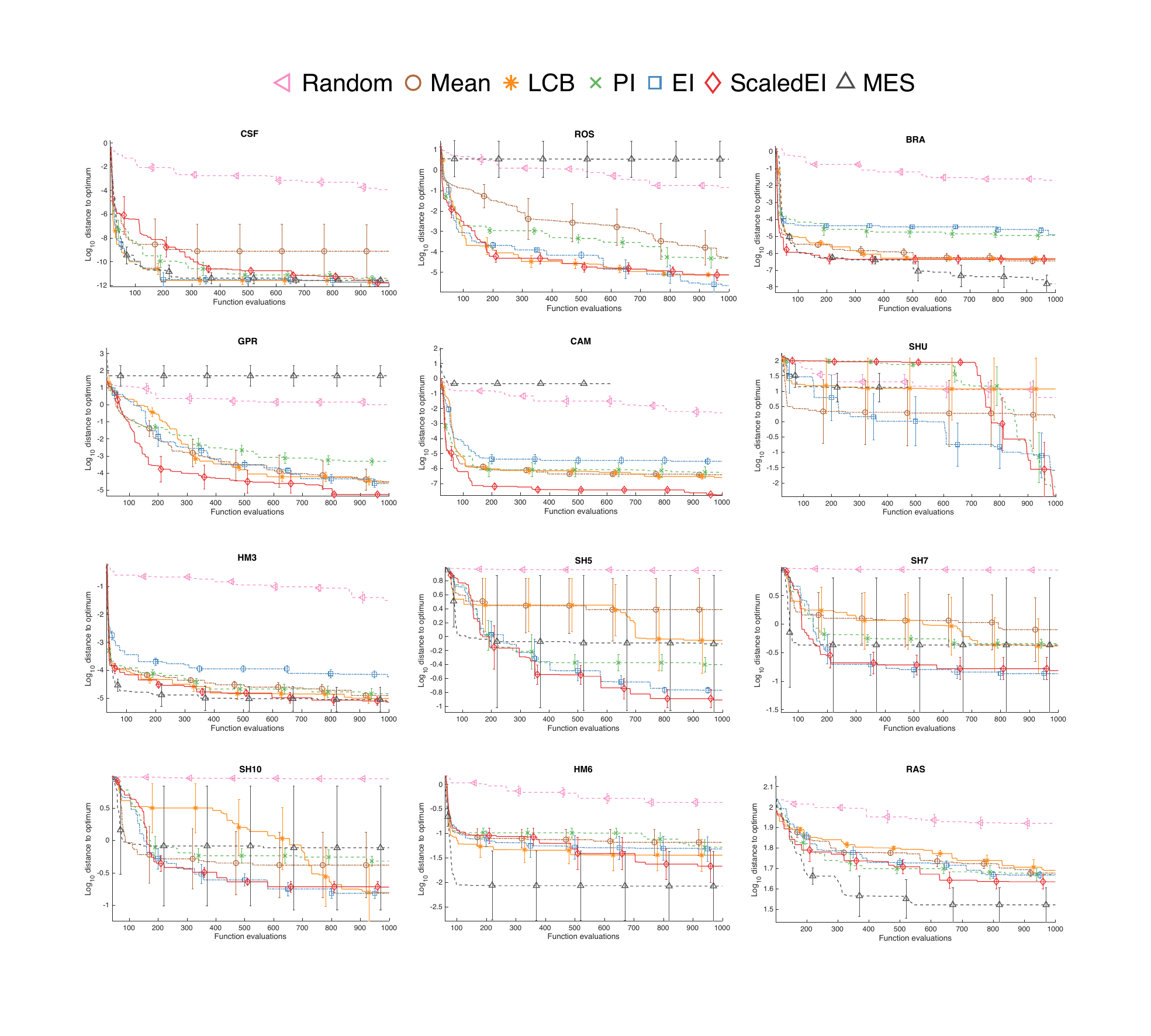}
\par\end{centering}
\caption{\textbf{Comparison of the log}\protect\textsubscript{\textbf{10}}\textbf{
distances (in function space) to the global optimum.} Each panel represents
a given benchmark function. The traces show the average distance over
the five design instantiations for the whole spectrum of iterations,
while the error bars show plus or minus one standard error of the
mean. These results use the ARD Mat\'{e}rn 5/2 kernel.\label{fig:matern_logdistances}}
\end{figure}

This section presents similar results to Section \ref{sec:benchmark-results},
but using a Gaussian process with the ARD Mat\'{e}rn 5/2 kernel. The only
exception is represented by the MES policy, whose code, provided by
\citet{Wang2017}, only allows for the ARD Squared Exponential kernel.
Figure \ref{fig:matern_logdistances} shows the full
spectrum of log\textsubscript{10} distances (in function space) to
the global optimum, for all function evaluations ranging from $n=100$
to $1000$. 

Tables \ref{tab:t_test_1000_matern52}, \ref{tab:t_test_600_matern52}
and \ref{tab:t_test_200_matern52} test the significance of the difference
in means of the log\textsubscript{10} distances at $n=1000,600,200$
respectively, for ScaledEI vs each of the remaining acquisition functions,
using a paired t-test with significance level 0.05. 
Table \ref{tab:t_test_1000_matern52}
shows that ScaledEI always outperforms the simple RND policy, and
often performs as well as (50-83\%) or significantly better (17-50\%)
than the competing algorithms, with only one exception. ScaledEI is
half of the time better than PI, followed by the information theoretic
competitor MES, where ScaledEI performs significantly better 33\%
of the times. Then follow LCB and EI, both outperformed 25\% of the
times and finally the conservative MN policy, outperformed only 17\%
of the times. 
Due to the information loss inherent in reducing an entire graph to a single number, the huge variations in the results of MN get lost, but they are clear in Figure \ref{fig:matern_logdistances}.
By chance a point in the initial Latin hypercube design can be near
the global optimum, and this will be fine-tuned because of the excessive
exploitative strategy. ScaledEI appears to be worse than EI in only
one test function. By inspecting in Figure \ref{fig:matern_logdistances}
the trace of the log\textsubscript{10} distance for that function,
SH10, we see that both algorithms are among the best performing methods,
and the difference, even if significant, is marginal in absolute terms.
Similar conclusions can be obtained from Tables \ref{tab:t_test_600_matern52}
and \ref{tab:t_test_200_matern52}.

We now compare Table \ref{tab:t_test_1000_matern52}, obtained using the ARD Mat\'{e}rn 5/2 kernel and Table \ref{tab:t_test_1000_se}, for the ARD Squared Exponential. 
ScaledEI performed always significantly better than the RND policy. Using the Squared Exponential kernel, ScaledEI is only once significantly better than MN while this happens twice using the Mat\'{e}rn kernel.
For the Squared Exponential kernel, the ScaledEI method has a significantly better performance in only one of the benchmark functions, compared to LCB, while in the Mat\'{e}rn one this happens three times. 
Using  the Mat\'{e}rn kernel, half of the time ScaledEI is better than PI, while using the infinitely-differentiable kernel, this happens only one third of the times. 
The comparison with EI is of particular interest. The column of t-test results are the same, apart from one function: SH10. Using the Squared Exponential kernel, there is not significant difference between ScaledEI and EI, while for the Mat\'{e}rn one the conclusion is that ScaledEI performed significantly worse. For both kernels, ScaledEI is significantly better than EI 25\% of the times. 
We recall that the code of MES is only available for the Squared Exponential kernel. ScaledEI using either a Squared Exponential or Mat\'{e}rn 5/2 kernel is 33\% of the times significantly better then MES with Squared Exponential kernel, and the remaining 67\% of the times they are not significantly different.

Overall, 81\% of the hypothesis test labels in Tables \ref{tab:t_test_1000_se} to \ref{tab:t_test_200_se} versus Tables \ref{tab:t_test_1000_matern52} to \ref{tab:t_test_200_matern52} agree between the two kernels, while in 19\% of the cases they are different. This suggests that the two kernels lead to similar results.

\begin{table}[tb]
\begin{centering}
\caption{\textbf{Statistical test for the significance in the mean difference
of the final log}\protect\textsubscript{\textbf{10}}\textbf{ distances.
}The ScaledEI acquisition function was tested against all remaining
acquisition functions using a paired t-test with significance level
0.05. Codes: 0 indicates a non significant difference and 1 (-1) indicates
that ScaledEI performed better (worse), i.e. it has a significantly
lower (higher) average distance. These results use the ARD Mat\'{e}rn
5/2 kernel.\label{tab:t_test_1000_matern52}}
\par\end{centering}
\centering{}%
\begin{tabular}{lcccccc}
\hline 
 & \multicolumn{6}{c}{ScaledEI vs}\tabularnewline
Test function  & RND  & MN  & LCB  & PI  & EI  & MES\tabularnewline
\hline 
CSF  & 1  & 0  & 0  & 0  & 0  & 0\tabularnewline
ROS  & 1  & 0  & 0  & 0  & 0  & 1\tabularnewline
BRA  & 1  & 0  & 0  & 1  & 1  & 0\tabularnewline
GPR  & 1  & 0  & 1  & 1  & 0  & 1\tabularnewline
CAM  & 1  & 1  & 1  & 1  & 1  & 1\tabularnewline
SHU  & 1  & 0  & 1  & 0  & 0  & 1\tabularnewline
HM3  & 1  & 0  & 0  & 0  & 1  & 0\tabularnewline
SH5  & 1  & 1  & 0  & 1  & 0  & 0\tabularnewline
SH7  & 1  & 0  & 0  & 1  & 0  & 0\tabularnewline
SH10  & 1  & 0  & 0  & 1  & -1  & 0\tabularnewline
HM6  & 1  & 0  & 0  & 0  & 0  & 0\tabularnewline
RAS  & 1  & 0  & 0  & 0  & 0  & 0\tabularnewline
\hline 
Same  & 0\%  & 83\%  & 75\%  & 50\%  & 67\%  & 67\%\tabularnewline
Better  & 100\%  & 17\%  & 25\%  & 50\%  & 25\%  & 33\%\tabularnewline
Worse  & 0\%  & 0\%  & 0\%  & 0\%  & 8\%  & 0\%\tabularnewline
\hline 
\end{tabular}
\end{table}

\begin{table}[tb]
\begin{centering}
\caption{\textbf{Statistical test for the significance in the mean difference
of the log}\protect\textsubscript{\textbf{10}}\textbf{ distances
at $\boldsymbol{n=600}$. }The ScaledEI acquisition function was tested
against all remaining acquisition functions using a paired t-test
with significance level 0.05. Codes: 0 indicates a non significant
difference and 1 (-1) indicates that ScaledEI performed better (worse),
i.e. it has a significantly lower (higher) average distance. These
results use the ARD Mat\'{e}rn 5/2 kernel.\label{tab:t_test_600_matern52}}
\par\end{centering}
\centering{}%
\begin{tabular}{lcccccc}
\hline 
 & \multicolumn{6}{c}{ScaledEI vs}\tabularnewline
Test function  & RND  & MN  & LCB  & PI  & EI  & MES\tabularnewline
\hline 
 CSF  & 1 & 0 & 0 & 0 & 0 & 0\tabularnewline
 ROS  & 1 & 0 & 0 & 1 & 0 & 1\tabularnewline
 BRA  & 1 & 0 & 0 & 1 & 1 & 0\tabularnewline
 GPR  & 1 & 0 & 0 & 0 & 0 & 1\tabularnewline
 CAM  & 1 & 1 & 1 & 1 & 1 & 1\tabularnewline
 SHU  & -1 & 0 & 0 & 0 & 0 & 0\tabularnewline
HM3 & 1 & 0 & 0 & 0 & 1 & 0\tabularnewline
SH5 & 1 & 0 & 0 & 0 & 0 & 0\tabularnewline
SH7 & 1 & 0 & 0 & 0 & 0 & 0\tabularnewline
SH10 & 1 & 0 & 0 & 1 & 0 & 0\tabularnewline
HM6 & 1 & 0 & 0 & 0 & 0 & 0\tabularnewline
 RAS  & 1 & 0 & 0 & 0 & 0 & 0\tabularnewline
\hline 
 Same  & 0\% & 92\% & 92\% & 67\% & 75\% & 75\%\tabularnewline
 Better  & 92\% & 8\% & 8\% & 33\% & 25\% & 25\%\tabularnewline
 Worse  & 8\% & 0\% & 0\% & 0\% & 0\% & 0\%\tabularnewline
\hline 
\end{tabular}
\end{table}
 
\begin{table}[tb]
\begin{centering}
\caption{\textbf{Statistical test for the significance in the mean difference
of the log}\protect\textsubscript{\textbf{10}}\textbf{ distances
at $\boldsymbol{n=200}$. }The ScaledEI acquisition function was tested
against all remaining acquisition functions using a paired t-test
with significance level 0.05. Codes: 0 indicates a non significant
difference and 1 (-1) indicates that ScaledEI performed better (worse),
i.e. it has a significantly lower (higher) average distance. These
results use the ARD Mat\'{e}rn 5/2 kernel.\label{tab:t_test_200_matern52}}
\par\end{centering}
\centering{}%
\begin{tabular}{lcccccc}
\hline 
 & \multicolumn{6}{c}{ScaledEI vs}\tabularnewline
Test function  & RND  & MN  & LCB  & PI  & EI  & MES\tabularnewline
\hline 
 CSF  & 1 & 0 & -1 & 0 & -1 & -1\tabularnewline
 ROS  & 1 & 1 & 0 & 1 & 0 & 1\tabularnewline
 BRA  & 1 & 1 & 1 & 1 & 1 & 0\tabularnewline
 GPR  & 1 & 0 & 1 & 0 & 0 & 1\tabularnewline
 CAM  & 1 & 1 & 1 & 1 & 1 & 1\tabularnewline
 SHU  & 0 & 0 & 0 & 0 & 0 & 0\tabularnewline
HM3 & 1 & 0 & 0 & 0 & 1 & 0\tabularnewline
SH5 & 1 & 0 & 0 & 0 & 0 & 0\tabularnewline
SH7 & 1 & 0 & 0 & 0 & 0 & 0\tabularnewline
SH10 & 1 & 0 & 0 & 0 & 0 & 0\tabularnewline
HM6 & 1 & 0 & 0 & 0 & 0 & 0\tabularnewline
 RAS  & 1 & 0 & 0 & 0 & 0 & 0\tabularnewline
\hline 
 Same  & 8\% & 75\% & 67\% & 75\% & 67\% & 67\%\tabularnewline
 Better  & 92\% & 25\% & 25\% & 25\% & 25\% & 25\%\tabularnewline
 Worse  & 0\% & 0\% & 8\% & 0\% & 8\% & 8\%\tabularnewline
\hline 
\end{tabular}
\end{table}



\end{document}